\documentclass{article}

\usepackage{PRIMEarxiv}
\usepackage{pdflscape}
\usepackage[utf8]{inputenc} 
\usepackage[T1]{fontenc}    

\usepackage{url}            
\usepackage{booktabs}       
\usepackage{amsfonts}       
\usepackage{nicefrac}       
\usepackage{microtype}      
\usepackage{lipsum}
\usepackage{caption}
\usepackage{fancyhdr}       
\usepackage{wrapfig}
\usepackage{graphicx}       
\usepackage{pifont}
\usepackage{array}
\graphicspath{{media/}}     

\usepackage[
backend=biber,
style=apa,
]{biblatex}
\usepackage{color}
\definecolor{blue}{RGB}{90,90,255}

\title{The Animal-AI Environment: A Virtual Laboratory For Comparative Cognition and Artificial Intelligence Research
\thanks{
Corresponding Author: Konstantinos Voudouris (kv301@srcf.net)}
}

\author{
  Konstantinos Voudouris, Ben Slater, Lucy G. Cheke \\
  Department of Psychology \\
  University of Cambridge \\
  Cambridge, UK\\
   \And
  Wout Schellaert, Jos\'{e} Hern\'{a}ndez-Orallo \\
  VRAIN / ValGRAI \\
  Universitat Politècnica de València \\
  València, Spain\\
    \And
  Marta Halina \\
  Department of History and Philosophy of Science \\
  University of Cambridge \\
  Cambridge, UK \\
 \And
  Matishalin Patel \\
  Centre for Data Science AI and Modelling \\
  University of Hull \\
  Hull, UK\\
  \And
  Ibrahim Alhas, Matteo G. Mecattaf, John Burden, Joel Holmes, Niharika Chaubey \\
  Leverhulme Centre for the Future of Intelligence  \\
  University of Cambridge \\
  Cambridge, UK\\
     \And
  Niall Donnelly \\
  Faculty of Environment, Science and Economy  \\
  University of Exeter \\
  Exeter, UK\\
   \And
   Matthew Crosby \\
   Department of Computing \\
   Imperial College London \\
   London, UK \\
}

\usepackage{setspace}
\usepackage[hidelinks]{hyperref}       
\begin{document}
\maketitle

\begin{abstract}

The Animal-AI Environment is a unique game-based research platform designed to facilitate collaboration between the artificial intelligence and comparative cognition research communities. In this paper, we present the latest version of the Animal-AI Environment, outlining several major features that make the game more engaging for humans and more complex for AI systems. These features include interactive buttons, reward dispensers, and player notifications, as well as an overhaul of the environment's graphics and processing for significant improvements in agent training time and quality of the human player experience. We provide detailed guidance on how to build computational and behavioural experiments with the Animal-AI Environment. We present results from a series of agents, including the state-of-the-art deep reinforcement learning agent \textit{Dreamer-v3}, on newly designed tests and the Animal-AI Testbed of 900 tasks inspired by research in the field of comparative cognition. The Animal-AI Environment offers a new approach for modelling cognition in humans and non-human animals, and for building biologically inspired artificial intelligence.
\end{abstract}

\keywords{Comparative Cognition \and Artificial Intelligence \and Cognition  \and Developmental Psychology \and Reinforcement Learning \and Computational Neuroscience}

\section{Introduction}

The Animal-AI Environment is the realisation of the idea that artificial intelligence (AI) research and cognitive science can and should work together to answer questions about cognition, behaviour, and intelligence. The Animal-AI Environment was first released in 2019 (\cite{beyret2019animalai, crosby2019animal}), with a suite of (initially) held-out cognitive tasks (The \textit{Animal-AI Testbed}) that were the basis of the \textit{The Animal-AI Olympics} competition in 2020 (\cite{crosby2020animal}). The environment is designed to offer a physically realistic \textit{virtual laboratory} for building experiments inspired by research in comparative cognition, the field that studies and compares behaviour in non-human animals. In this paper, we present a major update to the environment, facilitating the development of a much wider range of behavioural experiments than was previously possible. The newest version of the Animal-AI Environment will interest researchers working on the development and evaluation of cognitively inspired AI, as well as comparative and developmental psychologists working on developing computational models of behaviour in humans and non-human animals.

The Animal-AI Environment has a simple structure, designed to emulate the classical set-up of psychological research on non-human animals. The agent is spawned in a large arena, emulating an enclosure or laboratory, and it can be controlled by either a computational model or a human player. The objective is, within a specified time limit, to increase overall reward by obtaining rewarding objects and avoiding punishing ones. Rewarding objects correspond to appetitive stimuli such as food, while punishing objects correspond to aversive stimuli, such as electric shocks. The arena can be populated with a variety of other objects, such as ramps, blocks, tunnels, movable boxes, buttons, decoy goals, and dispensers. By using these objects as building blocks, an enormous variety of tasks can be generated, including setups similar to experiments from animal behaviour research (\cite{crosby2020animal, voudouris2022evaluating}). Thousands of distinct tasks can be easily procedurally generated to build comprehensive test batteries (\cite{crosby2020building}). The Animal-AI Environment is accompanied by several thousand open-source implementations of common experimental designs used in animal behaviour research, which can be freely used and adapted.

This article is divided into discrete sections; each is self-contained and can be read independently depending on the reader's interest.
In Section \ref{sec:virtuesofaai} we present several advantages of using the Animal-AI Environment for researchers in AI, comparative cognition, and those working at the intersection of these fields. We place the Animal-AI Environment in the context of other environments and initiatives in these fields, arguing that the Animal-AI Environment is unique in being the only initiative that serves both these communities at once, encouraging close collaboration. In Section \ref{sec:The New and Improved Animal-AI Environment} we present a full reference for the structure of the environment, including details of its features and how to run computational and behavioural experiments in the environment. In Section \ref{sec:battery-agents}, we outline the key agents that are implemented in the environment, serving as a starting point for research in AI and comparative cognition research. Finally, we present a short series of experiments to showcase the environment, these agents, and their potential for collaborative research.

\section{Collaborative Interdisciplinary Research with Animal-AI}\label{sec:virtuesofaai}

The Animal-AI Environment has been developed to facilitate collaborative and interdisciplinary research between artificial intelligence (AI) and comparative cognition. Comparative cognition offers the field of AI a wealth of experimental materials for studying and measuring intelligence in non-human systems (\cite{crosby2020building,hernandez2017measure}), providing inspiration for the development of better, more capable systems. Meanwhile AI research offers comparative cognition the ability to precisely model learning and behaviour computationally, in physically realistic environments that emulate laboratory conditions. Computational modelling of human behaviour has already significantly improved our understanding of human cognition and behaviour (\cite{griffiths_manifesto_2015, guest2021computational, zuidema2020five}), but computational modelling of non-human animal cognition is relatively nascent (\cite{voudouris2024computationalcomparative}).\footnote{See \cite{brea2023computational, baum2022yoking} for recent examples of this.} Collaboration between AI and comparative cognition can not only improve how we study the capabilities of AI systems, but also render more precise the theories and hypotheses we develop about animal behaviour (\cite{voudouris2024computationalcomparative}).

\begin{figure}[ht]
\centering
    \includegraphics[width=0.5\linewidth]{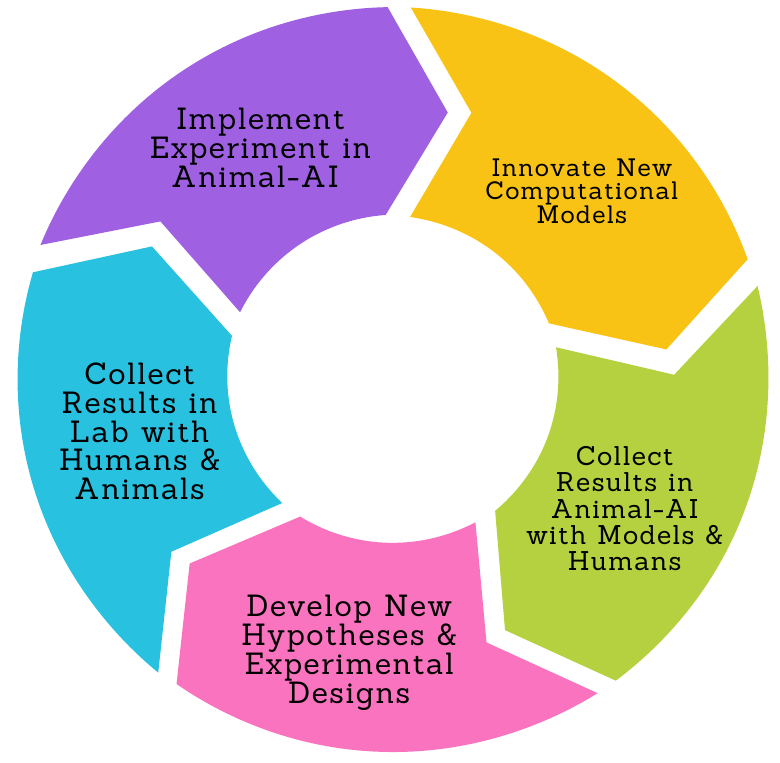}
    \caption{The Animal-AI Environment is a research tool that facilitates a virtuous cycle between AI research and animal behaviour research.}
    \label{fig:virtuous-cycle}
\end{figure}

A fundamental roadblock is the lack of a common platform for conducting comparable research on non-human animals and computational models. The Animal-AI Environment aims to overcome this obstacle. Primarily, the Animal-AI Environment is designed for building tests of spatial and physical cognition, including goal-directed navigation, object permanence, and tool use (\cite{crosby2019animal}). Physical cognition can be defined as any behaviour involving the manipulation of a non-social physical environment (\cite{lake2017building}). The agent in the Animal-AI Environment is therefore designed to take the place of an animal inside the laboratory, interacting with its environment in return for rewards while avoiding punishments, such as aversive stimuli.

Having a virtual laboratory for conducting behavioural tests on computational models encourages a virtuous cycle between AI and comparative cognition. Figure \ref{fig:virtuous-cycle} presents a simplified schematic of this cycle. Researchers start by replicating a comparative cognition experiment in the Animal-AI Environment, ranging from tests of instrumental learning to tests of tool use and causal reasoning. They can then evaluate the performance of computational models on that test, either the baseline systems that we provide or those they design through the standardised Python interface. How these models perform on realistic, lab-like tasks informs researchers of their flaws or merits, encouraging the development of new hypotheses, theories, and experimental designs which can then be developed in the physical laboratory with non-human animals. These new designs can then be re-implemented in the Animal-AI Environment. Not only can this process improve and render more precise the theories and models used to explain animal cognition, but it can also inspire improvements in the architectures and methods of AI (\cite{freire2024sequential,mitchener2022detect}).

\begin{figure}[ht]
\centering
    \includegraphics[width=0.9\linewidth]{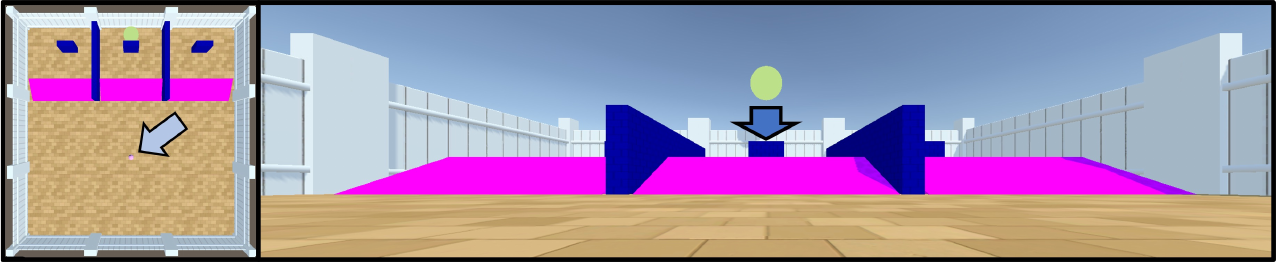}
    \caption{An example of a task testing object permanence in the Animal-AI Environment. \textbf{Left}: Bird's-eye view of the arena, with the location of the agent indicated by the grey arrow. \textbf{Right}: The agent's view from its starting position. It is frozen while the rewarding green sphere drops down behind the blue wall (see \cite{voudouris2022evaluating,voudouris2024investigating}).}
    \label{fig:3-cup-task}
\end{figure}

Figure \ref{fig:3-cup-task} presents an example of how Animal-AI can be the starting point for this interdisciplinary research. This experiment tests object permanence, the capacity to track occluded objects. It is a replication of classic tests of object permanence. A rewarding object is hidden in a cup, and the subject must correctly identify its location. Versions of this task have been conducted with infants (\cite{piaget1923origins, piaget1969intellectual}), several species of primate (e.g., \cite{herrmann2007humans}), and four species of parrot (\cite{krasheninnikova2019primate}). Comparative cognition has innovated several paradigms for detecting whether non-human animals are capable of object permanence, and these paradigms are immediately applicable to detecting whether embodied AI systems are capable of it too (\cite{voudouris2022evaluating}). However, there remains significant controversy on \textit{how} non-human animals (and infants) track occluded objects (see \cite{bremner2015perception,jaakkola2014animals,scholl2007object}). As \citeauthor{mareschal2000object} (\citeyear{mareschal2000object}) argues, computational modelling can help to elucidate the mechanisms underpinning object permanence. Indeed, in recent years, significant progress has been made towards AI systems that can track and reason about occluded objects (\cite{humantimescaleadaptation2022,piloto2022intuitive,tokmakov2021learning}). The Animal-AI Environment allows the evaluation of these sophisticated computational models on physically realistic experiments drawn from comparative cognition research, encouraging progress towards better theories about how animals track occluded objects \textit{and} towards building AI systems that reliably do it too.

To facilitate collaborative and interdisciplinary research, the Animal-AI Environment has the following design principles:
\begin{enumerate}
    \item A three-dimensional environment, with accurate and fast physics.
    \item Straightforward development of custom experiments, as well as procedural generation of variations of them.
    \item Minimal hardware requirements, so that behavioural experiments can be run locally or in the browser.
    \item Well-defined, functionally-specified objects for recreating existing experiments of physical cognition and for innovating new ones.
    \item Self-contained and relatively accessible to cognitive scientists without significant training in computer science or software engineering.
    \item Integration with existing frameworks for computational modelling and reinforcement learning, such as \texttt{Gymnasium} (\cite{towers_gymnasium_2023}) and \texttt{stable-baselines-3} (\cite{stable-baselines3}).
    \item Simple action space, to reduce the computational complexity of the learning problem, and to facilitate quick training and ease-of-use with adults, children, and non-human animals (such as primates).
    \item Appealing to play as a participant, facilitating collection of large and detailed human behavioural datasets.
    \item Openly available and free to use.
\end{enumerate}

The Animal-AI Environment is unique in its mission to serve interdisciplinary research between comparative cognition and AI. For example, while \citeauthor{allritz2022chimpanzees} (\citeyear{allritz2022chimpanzees}; see also \cite{miscov2024virtually}) present an example of a virtual environment, \textit{AP Explorer 3D}, for collecting data from primates on navigation and short-term memory tasks, they do not attempt to test computational models with this environment. Indeed, navigable virtual environments (i.e., open-world first-person video games) like the Animal-AI Environment have only very recently started to be used in the human cognitive sciences. These types of games, where players are embodied in a virtual avatar that can take several different actions and interact with a complex environment, offer numerous advantages such as increasing ecological validity by more closely approximating real world environments than traditional experiments (\cite{allen_using_2023}; \cite{allritz2022chimpanzees}). Video games facilitate more complex and nuanced experimental designs, as well as tapping into the popularity of computer games. In recent years, video games have been successfully used both for collecting rich human behavioural data and for computationally modelling that data (e.g., \cite{ansarinia2022cogenv, brandle_empowerment_2023, brown_crowdsourcing_2014, coutrot_global_2018,coughlan_toward_2019, coutrot_entropy_2022, dubey2018investigating,leduc-mcniven_serious_2018, mcnab_age-related_2015, pouncy_inductive_2022,rafner_digital_2022, schulz_exploring_2019, stafford_tracing_2014,voiskounsky_creativity_2017}; see \cite{allen_using_2023} for a review). The Animal-AI Environment offers a new opportunity to make similar progress at the intersection of comparative cognition and AI.

In contrast to comparative cognition and the cognitive sciences, there are several environments similar to the Animal-AI Environment that have been developed in the AI community. Most similar to the Animal-AI Environment in being three-dimensional environments with realistic physics are MuJoCo (\cite{todorov_2012}), DeepMind Lab (\cite{beattie2016deepmind,leibo2018psychlab}), Minecraft-based environments (Malm\"{o} \cite{johnson2016malmo}, MineRL \cite{guss2021minerl}, MineDoJo \cite{fan2022minedojo}), MiniWorld (\cite{chevalier2024minigrid}), XLand (\cite{humantimescaleadaptation2022}), Avalon (\cite{albrecht2022avalon}), and ThreeDWorld (\cite{gan2020threedworld,gan_open_2021}). The Animal-AI Environment distinguishes itself from these environments in being the only one targeted directly at the intersection between comparative cognition and AI, as well as being freely available, compatible with contemporary machine learning libraries in Python, facilitating customisable experimental designs, and being accompanied by an existing library of several thousand implementations of classic comparative cognition experiments (see Table \ref{tab:competitor-overview}). Similarly to XLand and ThreeDWorld, the Animal-AI Environment is built on top of Unity's physics engine (\cite{juliani2018unity}), providing fast real-world physics and three-dimensional graphics (\cite{ward2020using}). We also make use of Unity's open-source \texttt{ml-agents} Python API (\cite{juliani2018unity}), which is integrated with existing reinforcement learning frameworks, such as \texttt{Gymnasium} (\cite{towers_gymnasium_2023}), \texttt{Dopamine} (\cite{castro2018dopamine}), and \texttt{stable-baselines-3} (\cite{stable-baselines3}). Unlike these environments and similarly rich environments such as Avalon, the Animal-AI Environment is more constrained in its scope. While these other environments are invaluable for the development of large, data-driven computational models, they often demand expensive hardware and large research teams. The Animal-AI Environment is compact and simple enough to be used by small teams of AI researchers and cognitive scientists alike, while also scaling well to larger experiments. Nevertheless, Animal-AI's physics and visuals are rich enough to facilitate complex experiments that are still too difficult for state-of-the-art systems (\cite{crosby2020animal,voudouris2022direct,voudouris2024investigating}; see also Section \ref{sec:aai-testbed-experiment}). 

\begin{table}
\centering
 \caption{A comparison between the Animal-AI Environment and other realistic 3D environments in AI and cognitive science. \textit{Note}: Maintenance details are correct at the time of publication, with sources presented in Appendix \ref{app:maintenance-status}.}
  \begin{tabular}{ p{0.25\linewidth}>
  {\centering}p{0.17\linewidth} >{\centering}p{0.08\linewidth} >{\centering}p{0.07\linewidth} >{\centering}p{0.18\linewidth} >
  {\centering}p{0.1\linewidth}}
    \toprule
    Name (First Published) &
    Community & Freely Available & Custom Designs & Animal Experiment Library 
    & Actively Maintained \tabularnewline
    \midrule
    MuJoCo (2012) 
    & Robotics & \ding{52} & \ding{55} & \ding{55} 
    & \ding{52} \tabularnewline
    DM Lab (2016) 
    & AI & \ding{52} & \ding{52} & \ding{55} 
    & \ding{55} \tabularnewline
    Malm\"{o} (2016) 
    & AI & \ding{52} & \ding{52}  & \ding{55} 
    & \ding{55}\tabularnewline
     MiniWorld (2018) 
    & AI & \ding{52} & \ding{52}  & \ding{55} 
    & \ding{52}\tabularnewline
     MineRL (2019) 
    & AI & \ding{52} & \ding{52}  & \ding{55} 
    & \ding{52}\tabularnewline
    XLand (2021) 
    & AI  & \ding{55} & \textbf{?} & \textbf{?} 
    & \textbf{?}\tabularnewline
    Avalon (2022) 
    & AI & \ding{52} & \ding{52}  & \ding{55} 
    & \ding{55}\tabularnewline
    MineDoJo (2022) 
    & AI & \ding{52} & \ding{52}  & \ding{55} 
    & \ding{55}\tabularnewline
    ThreeDWorld (2020) 
    & AI, Human Psych. & \ding{52} & \ding{52}  & \ding{55} 
    & \ding{52}\tabularnewline
    APExplorer 3D (2022) 
    & Animal Psych. & \ding{52} & \ding{55}  & \ding{52} 
    & \textbf{?} \tabularnewline
    \midrule
    \textbf{Animal-AI (2019)} 
    & \textbf{AI, Animal Psych.} & \ding{52} & \ding{52}  & \ding{52} 
    & \ding{52}\tabularnewline
    \bottomrule
  \end{tabular}
  \label{tab:competitor-overview}
\end{table}


Most similar to the Animal-AI Environment in scope and focus are DeepMind Lab, MiniWorld, and the Minecraft-based builds MineRL and Malm\"{o}. DeepMind Lab is an environment for training and testing autonomous agents on a range of tasks. It is designed to have convincing three-dimensional physics and simple visuals, similar to the Animal-AI Environment. It was designed to facilitate research on general and autonomous agents, and to build tests inspired by neuroscience. Indeed, DeepMind Lab has been used to study exploration and learning in young children, inspired by work in developmental psychology (\cite{kosoy2020exploring}), as well as to improve reinforcement learning architectures (\cite{leibo2018psychlab}). However, DeepMind Lab requires experiments to be designed in Lua, a language that is much less familiar to researchers than Python, which is how users interact with the Animal-AI Environment (see \cite{chevalier2024minigrid}). Minecraft-based builds are similar to DeepMind Lab and the Animal-AI Environment in having a sophisticated physics engine and a definable ontology (i.e., blocks of different kinds that can populate the world). MiniWorld offers a three-dimensional environment with realistic physics for building customisable experiments for reinforcement learning systems. However, none of these environments are directly targeting comparative cognition research. The ontology of the Animal-AI Environment has been designed by comparative cognitive scientists to offer objects that are directly relevant to the methodologies of non-human animal behaviour research, including buttons, ramps, decaying and ripening rewards, and tools.

The Animal-AI Environment is accompanied by a library of experiments adapted from the comparative cognition literature, including 900 tasks testing spatial reasoning, working memory, numerosity, and tool-use (\cite{crosby2020animal}), over 12,000 tasks testing object permanence (\cite{voudouris2022evaluating}), over 5,000 tasks testing object affordances (\cite{rutar4924246general}), and over 260,000 basic tasks for interacting with the environment. The environment has been used to develop more sophisticated reinforcement learning architectures (\cite{freire2024sequential,lehuger2021fixed, mitchener2022detect,zakharov2021episodic}) as well as make progress on computational models of learning in non-human animals (\cite{lind2021can, cardoso2023stimulus}). It also facilitates direct comparison to human participants, enabling the evaluation of the state-of-the-art in reinforcement learning (\cite{burden2023inferring,voudouris2022direct, voudouris2024investigating}) and multi-modal large language models (\cite{mecattafslater2024lessconversation}). The Animal-AI Environment is therefore at the foundation of a rich and active research community at the intersection of comparative cognition and AI.

\section{The Animal-AI Environment: A Comprehensive Overview}
\label{sec:The New and Improved Animal-AI Environment}

In this section, we provide a comprehensive outline of the Animal-AI Environment, highlighting the changes that have been made since earlier releases (\cite{beyret2019animalai,crosby2020animal}). 

\subsection{Environment Design and Modes of Use}\label{sec:env-design-and-spec}

The Animal-AI Environment has two principal modes of use:
\begin{itemize}
    \item \textbf{Play Mode}: A full visual display for humans, enabling interaction with simple keyboard controls.
    \item \textbf{Training Mode}: A lightweight Python API for training and testing computational agents on visual output from the Animal-AI Environment.
\end{itemize}

The Animal-AI Environment can be configured to create episodic tasks for humans and artificial agents. Each episode takes place inside an arena that can contain a wide range of different objects together with the agent. More information on all of these features and more is described in the following sections. As noted, the Animal-AI Environment is built on top of Unity's \texttt{ml-Agents} library (\cite{juliani2018unity}).\footnote{We provide an accompanying \texttt{animalai} Python package here: \href{https://pypi.org/project/animalai/}{https://pypi.org/project/animalai/}.}

Substantial graphical improvements have been made to the Animal-AI Environment---not only cosmetic changes to improve the environment's appearance (textured walls, animal skins, improved lighting) but also efforts to optimise environment rendering for faster performance. This includes pruning unnecessary casting and receiving of shadows; making use of the performance-orientated Universal Render Pipeline (URP);\footnote{See \href{https://unity.com/srp/universal-render-pipeline}{https://unity.com/srp/universal-render-pipeline}.} simplifying the shaders and lighting model used to render 3D objects; and reducing the resolution and anti-aliasing of textures and shadows. These optimisations complement efforts to create a more vibrant, engaging, and visually straightforward environmental appearance for human players, particularly children, while also increasing the frames per second (fps) at which the environment can run when using camera-based observations (see Section \ref{sec:RL-setup}).

\subsection{Environment Specification}

\subsubsection{The Arena \& Agent}

As in earlier versions of the environment, the arena is a 40$\times$40 unit space (where the agent is a sphere with a diameter of 1), including a tiled floor and four surrounding walls resembling white picket fences.\footnote{Note that there is no explicit height restriction on the arena.} 
The \texttt{Arena} object can be configured with three parameters. The arena time, \texttt{timeLimit}, provides the total number of time steps an agent or user can take before the episode ends. The reward for an episode begins at 0 and decreases in increments of $\frac{1}{t}$ per timestep, where \texttt{t} is the value of \texttt{timeLimit}, increasing when the agent interacts with a positively valenced object (see \ref{para:rewards}) and decreasing upon interaction with a negatively valenced object. If \texttt{timeLimit} is set to \texttt{0}, then the reward value no longer decreases with each time step. The arena pass mark, \texttt{passMark}, allows the researcher to define the reward threshold that should constitute a `pass' on that particular episode, providing an easy metric for success or failure across episodes. It also determines the notifications (if turned on) that players receive at the end of an episode (see Section \ref{sec:human-experiments}).

\begin{figure}[ht]
\centering
    \includegraphics[width=0.6\linewidth]{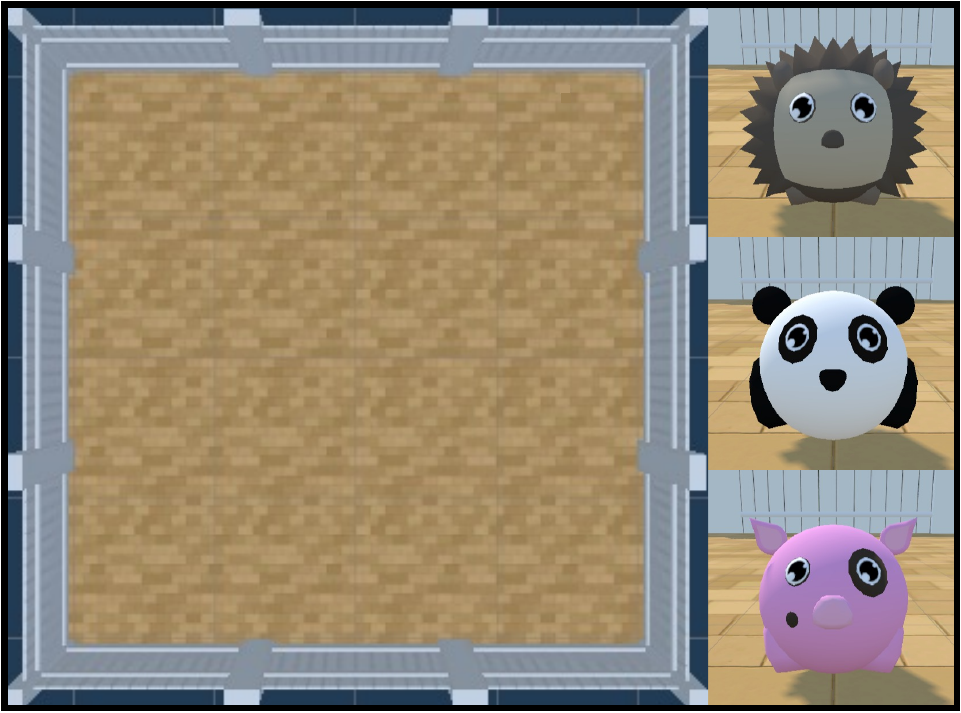}
    \caption{The \texttt{Arena} and the three \texttt{Agent} characters that can be selected for play and testing. \textbf{Top}: \texttt{"hedgehog"}. \textbf{Centre}: \texttt{"panda"}. \textbf{Bottom}: \texttt{"pig"}.}
    \label{fig:ArenaAndAgent}
\end{figure}

The agent is of size 1$\times$1$\times$1 and can spawn anywhere in the \texttt{Arena}. It can take one of three forms: a hedgehog, a panda, or a pig, enabling the creation of an engaging narrative for studies with children (see Figure \ref{fig:ArenaAndAgent}). The agent can be frozen at the start of an episode for a specified number of steps, using the \texttt{frozenAgentDelays} parameter, a new feature in the Animal-AI Environment. There is no reward decrement during the frozen period and all other objects continue to move and interact. This allows the agent to observe occurrences in the environment before being allowed to move.

To replicate the view of an animal in a laboratory, the agent observes the environment from a first-person perspective, both in ``Play Mode'' and ``Training Mode''. Two further perspectives are offered in ``Play Mode''. `Third-person view' follows the agent around from a perspective slightly above and behind the agent. `Birds-eye view' offers a view of the arena in its entirety from above. These perspectives are offered so that researchers can view their experimental designs from multiple perspectives. Researchers can specify whether players in ``Play Mode'' can change their camera angle, and which camera angle should be the default. In ``Training Mode'', the API only returns the `First-person view' to simulate the perspective of an animal in the laboratory. The number of pixels or raycasts (another form of visual observation; see Section \ref{sec:RL-setup}) in this image is configurable. 

The \texttt{Arena} object can be configured to have `lights out' periods, in which all pixels or raycasts in the `first-person view' are replaced with zeros. For a human player, this results in a black screen. Lights out periods can be specified for specific time step intervals, or on an alternating schedule.

\subsubsection{Immovable Objects}

As in earlier versions of the environment, there are five objects that cannot be moved by the agent: opaque and transparent tunnels and walls, and opaque ramps. All of these objects can be resized, rotated, and re-positioned freely by experiment designers via a configuration file. All opaque objects can be varied in colour in a 256$\times$256$\times$256 RGB colour space (see Figure \ref{fig:Immoveables}). The characteristics of these five immovable objects are:

\begin{figure}[h]
\centering
    \includegraphics[width=0.6\linewidth]{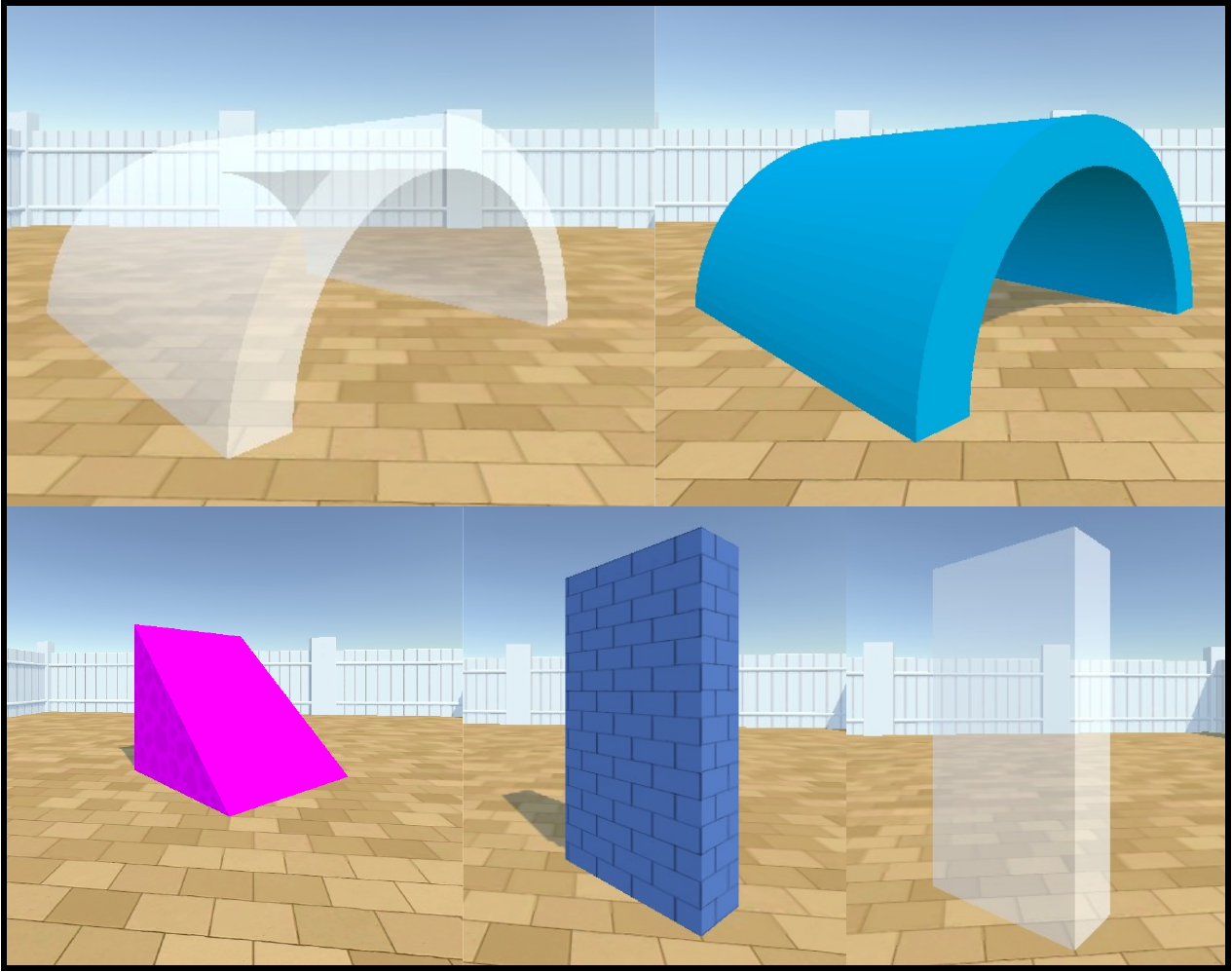}
    \caption{The Immovable objects in the Animal-AI Environment. \textbf{Top Left}:  \texttt{CylinderTunnelTransparent}. \textbf{Top Right}: \texttt{CylinderTunnel}. \textbf{Bottom Left}: \texttt{Ramp}. \textbf{Bottom Centre}: \texttt{Wall}. \textbf{Bottom Right}: \texttt{WallTransparent}.}
\label{fig:Immoveables}
\end{figure}

\begin{itemize}
\item Opaque Walls: An opaque cuboidal object used to construct obstacles to be navigated around. The colour of the walls can be changed. In the Animal-AI Environment, unlike in some earlier versions, opaque walls have a ``brickwork'' texture of black lines.
\item Transparent Walls: A non-opaque cuboidal object used to construct obstacles to navigate around.
\item Ramps: An opaque right-angled prism object. They can be climbed on/over, if the height-length ratio does not exceed 4:1. In the Animal-AI Environment, unlike in earlier versions, ramps have a patterned texture of lighter and darker colour.
\item Opaque Tunnels: An opaque hemi-cylindrical tunnel used as an obstacle to be navigated through and/or around.
\item Transparent Tunnels: A non-opaque hemicylindrical tunnel used as an obstacle to be navigated through and/or around.
\end{itemize}

\subsubsection{Movable Objects}

There are six movable objects in the Animal-AI Environment: hollow boxes, light and heavy rectangular blocks, U-shaped, L-shaped, and J-shaped blocks, all of which can be pushed around.
These blocks can be used as tools for moving other objects (other movable objects or any of the valenced rewards discussed below). These objects can be resized, rotated, and re-positioned freely, but their colour cannot be changed. In the Animal-AI Environment, unlike in earlier versions, blocks do not have a ``cardboard box''- or ``wood''- style texture, instead being uniform grey in colour with light grey trims along the vertices.

\begin{itemize}
\item Light Blocks: An opaque cuboidal object. Light blocks have a mass of 1.
\item Heavy Blocks: An opaque cuboidal object. These objects function the same as light blocks, except that heavy blocks have a mass of 2.
\item U-Shaped Blocks: An opaque object with a squared ``U'' shape that facilitates pushing and pulling of other movable objects. U-shaped blocks have a mass of 1.5.
\item L- and J-Shaped Blocks: Opaque objects with a squared ``L'' or ``J'' shape (mirror images of each other) that facilitate pushing of other objects. L- and J-shaped blocks have a mass of 1.5.
\item Hollow Boxes: Opaque cuboidal hollow objects with an opening at the top, allowing them to contain other objects. They have a mass of 1.5.
\end{itemize}

\begin{figure}[h]
\centering
    \includegraphics[width=0.9\linewidth]{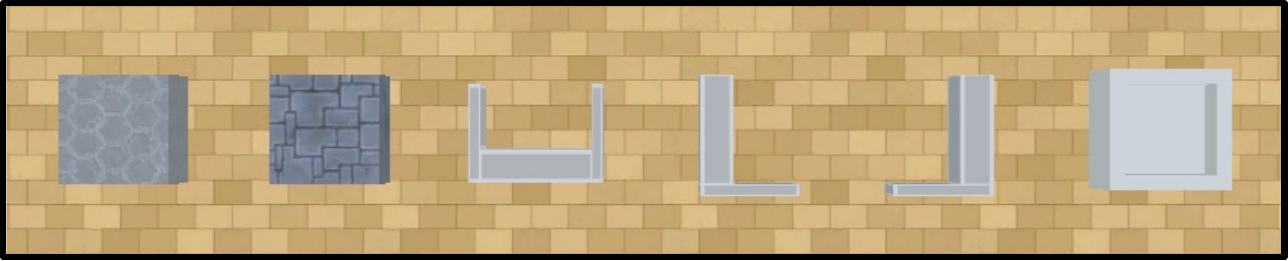}
    \caption{The movable objects in the Animal-AI Environment. From \textbf{Left} to \textbf{Right}: \texttt{LightBlock}, \texttt{HeavyBlock}, \texttt{UBlock}, \texttt{LBlock}, \texttt{JBlock}, \texttt{HollowBox}.}
\label{fig:Moveables}
\end{figure}

\subsubsection{Valenced Objects}\label{para:rewards}

\begin{figure}[h]
\centering
    \includegraphics[width=0.9\linewidth]{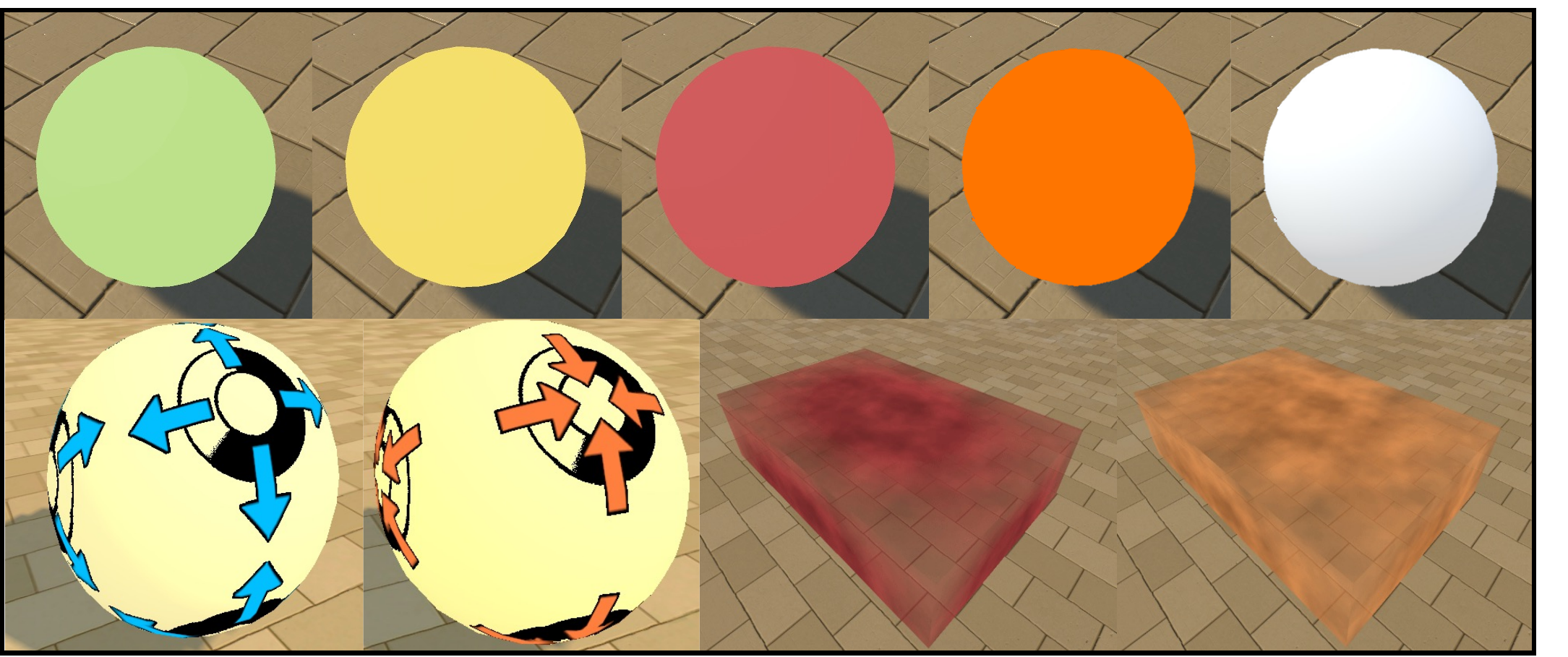}
    \caption{The Valenced Objects in the Animal-AI Environment. \textbf{Top Far Left}:  \texttt{GoodGoal}/\texttt{GoodGoalBounce}. \textbf{Top Centre Left}: \texttt{GoodGoalMulti}/\texttt{GoodGoalMultiBounce}/\texttt{GrowGoal}/\texttt{ShrinkGoal}. \textbf{Top Centre}: \texttt{BadGoal}/ \texttt{BadGoalBounce}. \textbf{Top Centre Right}: \texttt{BadGoalMulti}/\texttt{BadGoalMultiBounce}. \textbf{Top Far Right}: \texttt{DecoyGoal}/\texttt{DecoyGoalBounce} \textbf{Bottom Far Left}: \texttt{RipenGoal}. \textbf{Bottom Centre Left}: \texttt{DecayGoal}. \textbf{Bottom Centre Right}: \texttt{DeathZone}. \textbf{Bottom Far Right}: \texttt{HotZone}.}
\label{fig:Rewards}
\end{figure}

Upon making contact, positively and negatively valenced objects can increase or decrease reward respectively. These objects vary in both size, shape, and appearance. The colour of valenced objects can be configured by the researcher, with the default colours being those presented in Figure \ref{fig:Rewards}. Decaying, ripening, growing, and shrinking valenced objects are new to the Animal-AI Environment. In what follows, we refer to spherical valenced objects as ``goals'' and cuboidal regions of space that have valence as ``zones''.

\begin{itemize}
\item Non-Moving Goals (episode- and non-episode-ending): Coloured spheres with a valence proportional to their size. They only move under the force of gravity or collision. These objects can be positive or negative and episode ending in the cases of \texttt{GoodGoal} and \texttt{BadGoal}, positive or negative and non-episode ending in the cases of \texttt{GoodGoalMulti} and \texttt{BadGoalMulti}, or neutral and non-episode ending in the case of \texttt{DecoyGoal}. They are solid and opaque and their colours can be changed (with the defaults shown in Fig. \ref{fig:Rewards}).
\item  Moving Goals (episode- and non-episode-ending): Identical to non-moving goals, except that they have an initial velocity in a prespecified or randomised direction at the start of the episode. They are denoted with the suffix `\texttt{-Bounce}'. They are solid and opaque.
\item Growing and Shrinking Goals: Similar to ripening and decay goals respectively, except their physical size changes proportional to their valence. The start and end size can be configured. Growth/Shrinkage can be delayed for a specified number of steps. Collecting these goals does not end an episode. A growth goal will cease to grow if the space around it becomes restrictive. They are solid and opaque. 
\item Decaying and Ripening Goals: Similar to non-moving goals, except that their valence changes over time, while their physical size remains constant from spawn time. The valence decreases over time for decaying goals, and increases with ripening goals. The difference is indicated by six 2-dimensional glyphs on the goal's surface, as well as colour changes. Decaying goals change from yellow to grey, while ripening goals change from grey to yellow. Decaying goals have glyphs of arrows pointing inwards, with a large black ring that empties over time. Ripening goals have glyphs of arrows pointing outwards, with a ring that fills over time. Both goals can have a configurable ``full'' and ``empty'' valence and a configurable delay between their starting valence (full or empty) and their ultimate valence (empty or full). Collecting these goals does not end an episode. They are solid and opaque.
\item Death Zone: A translucent red zone. When the agent makes contact with a death zone, its reward decreases by 1 and the episode ends. In the Animal-AI Environment, unlike in earlier versions, death zones occupy a three-dimensional region of space, through which other objects, including the agent, can pass.
\item Hot Zone: A translucent orange zone. When the agent enters the hot zone, reward decreases 10$\times$ faster than with time alone. It does not end the episode upon contact. In the Animal-AI Environment, unlike in earlier versions, hot zones occupy a three-dimensional region of space, through which other objects, including the agent, can pass. If a hot zone and death zone overlap, the death zone takes precedence.
\end{itemize}

\subsubsection{Dispensers} 

\begin{figure}[h]
\centering
    \includegraphics[width=0.9\linewidth]{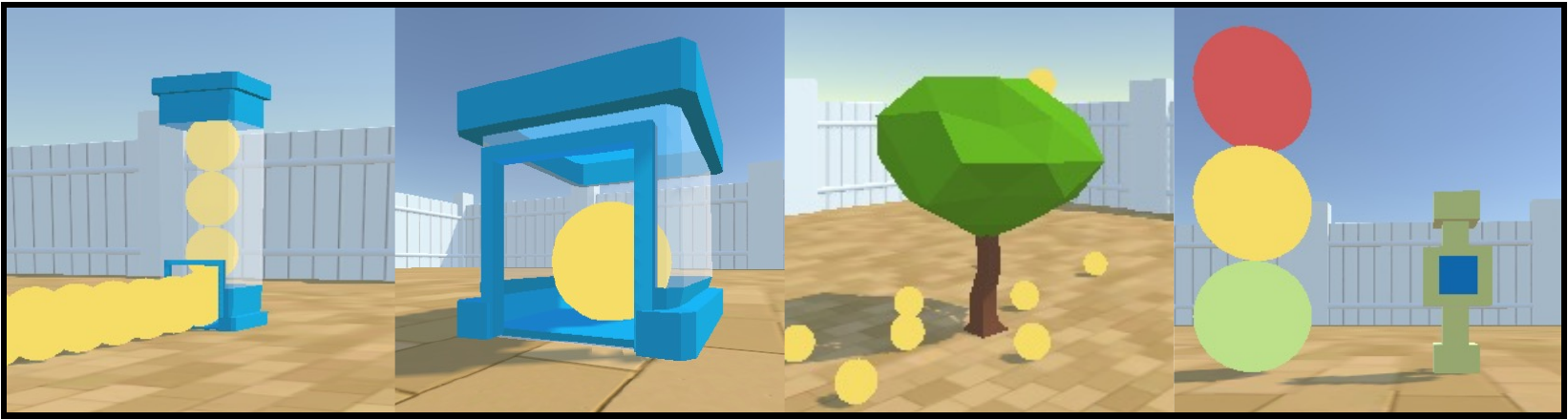}
    \caption{The dispenser objects in the Animal-AI Environment. From \textbf{Left} to \textbf{Right}:  \texttt{SpawnerDispenserTall} producing unlimited \texttt{GoodGoalMulti}, \texttt{SpawnerDispenserShort} producing unlimited \texttt{GoodGoalMulti}, \texttt{SpawnerTree} producing unlimited \texttt{GoodGoalMulti}, \texttt{SpawnerButton} producing unlimited \texttt{GoodGoal}, \texttt{GoodGoalMulti}, and \texttt{BadGoal} with equal probability.}
\label{fig:Dispensers}
\end{figure}

Dispensers are designed to emulate natural (e.g., tree spawners) and laboratory-like (e.g., goal dispensers) elements that dispense valenced objects, allowing replication of various human and non-human animal experiments. All of these objects are new to the Animal-AI Environment. See Figures \ref{fig:Dispensers}.

\begin{itemize}
\item Spawner Tree: A tree-shaped object that spawns \texttt{GoodGoalMulti} objects, which initially grow in the tree branches, then fall to the ground after a configurable time delay. The number of \texttt{GoodGoalMulti} spawned can be varied, and are dropped from the tree at random, in the style of naturally falling fruit. Tree Spawners have a fixed size and colour.
\item Tall Spawner Dispenser: An immovable, cuboidal object designed to look like a simplified vending-machine/food magazine that spawns \texttt{GoodGoalMulti} through a hatch. The colour of the dispenser, the number of \texttt{GoodGoalMulti} spawned, and the delay between spawning can be configured. The size of the dispenser is fixed.
\item Short Spawner Dispenser: A more compact version of the Goal Dispenser that also spawns \texttt{GoodGoalMulti} into the environment after a configurable time delay through a hatch. The size of the dispenser is fixed.
\item Spawner Button: An opaque, immovable object with fixed size and colour (beige pillar, blue button). The \texttt{SpawnerButton} object can spawn three types of valenced object, \texttt{GoodGoal}, \texttt{GoodGoalMulti}, and \texttt{BadGoal}. The probability of a goal spawning upon pressing the button can be defined. The probability of spawning each of these three goal types can also be weighted, to increase or decrease the likelihood of it appearing when the button is pressed. The position of where the goal is spawned can also be defined. Finally, the number of steps it takes to reset the button can be defined.
\end{itemize}

\subsubsection{Sign Boards}

\begin{figure}[h]
\centering
    \includegraphics[width=0.7\linewidth]{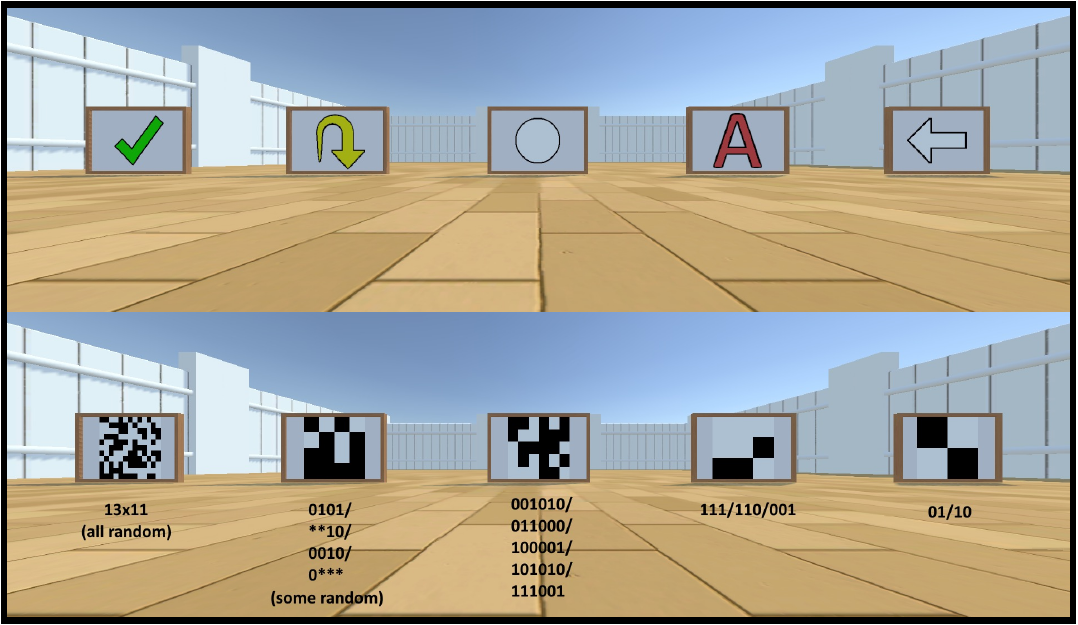}
    \caption{The Sign Board object in the Animal-AI Environment. \textbf{Top}:  \texttt{SignBoard}s with 5 preset symbols (\textbf{Left} to \textbf{Right}: \texttt{"tick"}, \texttt{"u-turn-arrow"}, \texttt{"circle"}, \texttt{"letter-a"}, \texttt{"left-arrow"}. \textbf{Bottom}: \texttt{SignBoard}s with defined and randomised black grids. The syntax for defining these is overlayed.}
\label{fig:signposters}
\end{figure}

Sign boards, a new addition to the Animal-AI Environment, enable the presentation of two-dimensional neutral stimuli to signal, for example, the presence or absence of valenced objects. They allow the replication of conditioning experiments common in cognitive science (see, e.g., \cite{cardoso2023stimulus}) as well as for general signposting (e.g. with arrows). See Figure \ref{fig:signposters}.

\begin{itemize}
    \item Sign board: An opaque, cuboidal, immovable object that displays a two-dimensional visual image. They have a fixed size 
    and a preset list of display symbols (including arrows, letters, and shapes), which can be used to non-linguistically signal events in the environment. Pixel grids can also be presented, with full or partial randomisation. Pixel grids, as well as the preset list of display symbols, can be given user-defined colours on the RGB spectrum.
\end{itemize}

\subsection{Physics Simulation}

The Animal-AI Environment evolves in discrete time. The state of the environment at a certain time step is the complete set of information about the objects and agent that exists within it. This encompasses a multitude of factors, including the position of objects, the agent's position and orientation, time step information, velocities, contact with objects, and many other relevant factors. At each step, the agent can take one of nine discrete actions, transitioning the environment to its next state, with the default action being \textit{No Action}:
\begin{itemize}
    \item \textbf{No Action}: the agent has no force acting on it in any direction, apart from the force of gravity or of other objects colliding with it.
    \item \textbf{Forwards}: the agent has force acting on it in the forwards direction (towards the observable part of the environment).
    \item \textbf{Left}: the agent rotates 6 degrees left (negative offset).
    \item \textbf{Right}: the agent rotates 6 degrees right (positive offset).
    \item \textbf{ForwardsLeft}: the agent has force acting on it in the forwards direction and rotates 6 degrees left.
    \item \textbf{ForwardsRight}: the agent has force acting on it in the forwards direction and rotates 6 degrees right.
    \item \textbf{Backwards}: the agent has force acting on it in the backwards direction (away from the observable part of the environment).
    \item \textbf{BackwardsLeft}: the agent has force acting on it in the backwards direction and rotates 6 degrees left.
    \item \textbf{BackwardsRight}: the agent has force acting on it in the backwards direction and rotates 6 degrees right.   
\end{itemize}

How the environment transitions from one state to the next is defined by the environment's transition function, and it depends on the preceding states of the environment and the action that the agent takes on that step. The Animal-AI Environment is built with Unity's physics engine, which has realistic Newtonian physics and Euclidean geometry that define the transition function. There are several forces that act on the objects in the environment that derive from the Unity physics engine:
\begin{itemize}
    \item \textbf{Gravitational Force}: Every object experiences a constant force downwards (on the y-axis), simulating gravity. 
    \item \textbf{Normal Force}: An object placed on a surface experiences an upward force equal and opposite to the gravitational force.
    \item \textbf{Applied Force}: When the agent moves forwards or backwards, or a \texttt{GoodGoalBounce}, \texttt{GoodGoalMultiBounce}, or a \texttt{BadGoalBounce} move in a direction, this is due to an applied force in that direction.\footnote{In the case of moving goals, this force only applies on the first time step.}
    \item \textbf{Reaction Force}: When objects collide into each other, the object being collided with receives a force in the opposite direction to the collision path, proportional to its mass.
    \item \textbf{Frictional Force}: There is a frictional force that opposes the direction of motion. All objects in the environment have high frictional force.
    \item \textbf{Drag Force}: Objects moving in the environment experience air resistance (drag) which opposes their movement proportional to their size.
    
\end{itemize}

\subsection{Reward Signal \& Health}

The value of the reward signal can be positive or negative. An agent accrues or loses reward through touching positively or negatively valenced objects. Additionally, the agent receives a minor negative reward after each step, which is equivalent to $-\frac{1}{t}$ per timestep, where $t$ refers to the number of timesteps in the episode. In a physical laboratory, food is often withheld from animals prior to an experiment so that they are motivated to obtain food as quickly as possible.\footnote{Note that this is not always the case. For instance, in contemporary great ape studies, highly desirable and not readily available rewards are used.} Including a reward decrement on each timestep emulates this necessity to obtain food in the laboratory. However, sometimes this design is not required. Therefore, when $t$ is 0, there is no time step limit, and reward does not decrement on each time step.

As an interpretable indicator of current reward for human players, we include a health indicator. This starts at $100$ and the agent fails the episode when it reaches $0$. The rate of health decrement is proportional to reward decrement. Health is replenished upon obtaining positively valenced objects proportional to the object's value, to a maximum value of $100$.

\subsection{Using Animal-AI with Computational Models}\label{sec:RL-setup}

The Animal-AI Environment has a straightforward Python API for integration with computational models, based on the \texttt{ml-agents} library (\cite{juliani2018unity}). The environment returns information about its state at each time step, including the agent's first-person observations, position, velocity, and current reward, and the model must return one action from the set of permissible actions in order to evolve the environment. Thus, any model implemented in Python that returns single actions in response to observations can be integrated into the Animal-AI Environment.

A popular class of computational models that take observations and return actions are reinforcement learning agents, based on algorithms that seek to find sequences of actions that optimise reward. Reinforcement learning agents can be implemented directly in the \texttt{ml-agents} Python API. Alternatively, the environment can be wrapped to form a \texttt{Gymnasium} environment (\cite{towers_gymnasium_2023}). This enables researchers to make use of the myriad of reinforcement learning agents that have been implemented elsewhere in the literature, as well as build new agents according to standardised interfaces which can interact both with the Animal-AI Environment and the many other similar environments that exist. 

Since the Animal-AI Environment is partially observable, the agent does not have access to complete information about the current state of the environment. The extent of the observability depends on the observation type. \textit{Camera} observations allow for the closest replication of animal behaviour in the laboratory, allowing the agent to observe an image of the environment in front of them. The resolution of the image can be set to $k \times k \times 3$ where $4 \leq k \leq 512$ for the RGB colour channels.  Grayscale can be turned on or off. With grayscale observations, the camera observation is simplified to a $k \times k$ matrix. To produce a simplified observation space, the agent can also make use of \textit{raycast} observations. The agent can emit `rays' along its $x$-$z$ (horizontal) plane which intersect objects and return information about what objects the rays hit and how far away they are. The first ray is emitted directly ahead, with further rays fanned out equally, starting on the left. The number of rays, and the angle over which they are distributed, can be specified. Raycast observations return a matrix of $8 \times r$ where $r$ is an odd number. Raycast observations allow for quicker learning and incur less computational overhead than camera observations. Finally, the agent can receive observations about the agent's current health, its velocity in ($x,y,z$), and its global position in ($x,y,z$). All of these observation types can be used in any combination.

\subsection{Using Animal-AI with Human Participants}\label{sec:human-experiments}

\begin{figure}[h]
    \centering
    \includegraphics[width=0.5\linewidth]{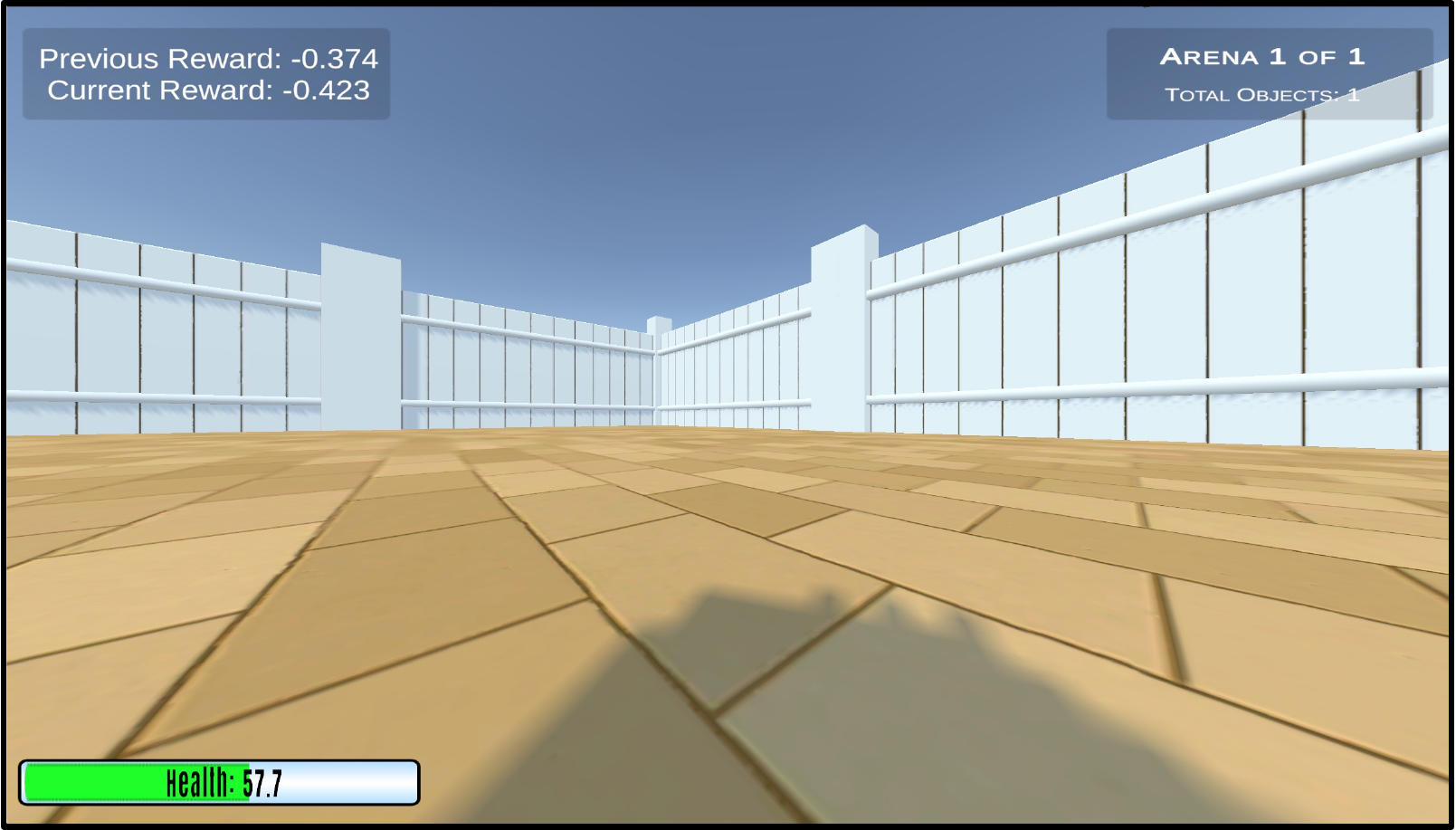}
    \caption{The first-person view of the playing participant, with the health bar in the bottom left corner, the current and previous episode reward in the top left corner, and the arena index in the top right corner.}
    \label{fig:ParticipantViews}
\end{figure}

Human behavioural testing is facilitated in the Animal-AI Environment using ``play mode''. The participant interacts with the environment using the arrow-keys, [W, A, S, D], or with a controller connected to the testing system. Participants are provided with a high-resolution 1200$\times$800 pixel display of the environment at 60fps, from one of three perspectives, with their `health' in the bottom-left corner. `Health', a new feature of the Animal-AI Environment, starts at 100 and steadily decreases with time. It is replenished when reward increases. Obtaining a reward of size 0.5 increases health by 50. Health cannot be over 100. The `Health' bar turns continuously from blue through green to orange and finally red as it runs from 100 to 0. Participants are provided with the current reward value and the reward value at the end of the previous episode in the top-left corner. At the end of an episode, researchers can choose whether the participant is presented with a notification of whether they passed or failed. See Figure \ref{fig:ParticipantViews}.

Researchers can collect information on participant location and reward at each time step of an episode, permitting analysis of not just whether a participant passed or failed a task, but the path they took across the episode. This gives an insight into \textit{how} a participant passed or failed. We provide full scripts in the supplementary material for conducting testing with humans.

\subsection{Designing Experiments in Animal-AI}
\label{sec:Designing Experiments in Animal-AI}

Building tasks in the Animal-AI Environment is straightforward. Researchers can build task configurations in an interpretable and easy to use \texttt{YAML} configuration file, specifying the position, size, colour, and special parameters of the objects required for testing. The Animal-AI Environment then interprets configuration files and renders those objects in a new instance of the environment. The agent is always spawned first in the environment, and takes precedence over all other objects, regardless of the order in which the agent is specified in the configuration file. Otherwise, priority is in the order of the objects specified in the configuration file. Several arena configurations can be specified in a single file, meaning that sequences of tests can be presented in the order specified by the researcher. Researchers can also create multi-arena configurations, where the agent moves from one arena to the next within the same episode. This enables the design of carefully crafted hierarchical training and testing curricula, where the difficulty of tests can be adjusted and prespecified.

We also provide a configuration design assistant, which allows users to visualise, edit, and create configuration files through a graphical web application written in Python.\footnote{See \url{https://github.com/Kinds-of-Intelligence-CFI/aai-config-assist}.} The assistant facilitates task building by removing the need to manually write \texttt{YAML} files and by giving users immediate visual feedback about the configuration being designed. The assistant provides considerable interactivity with many application callbacks. These include moving, resizing, deleting, and adding items to an Animal-AI arena as well as fully designing a new arena from scratch.

\begin{figure}[h]
\centering
    \includegraphics[width=0.9\linewidth]{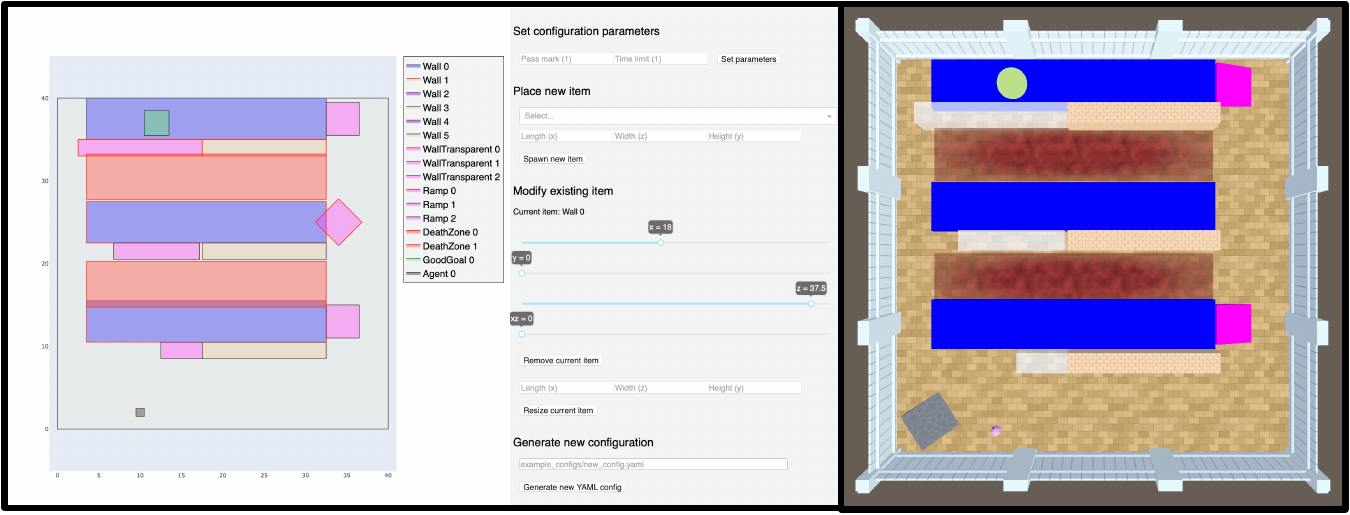}
    \caption{The Animal-AI configuration design companion tool. \textbf{Left}: The graphical user interface for the tool, built in Python. \textbf{Right}: The rendered configuration in the Animal-AI Environment. Note the displaced grey ramp in the bottom left of the arena, as a result of a detected clash in the configuration tool.}
\label{fig:configuration-tool}
\end{figure}

Figure \ref{fig:configuration-tool} shows the configuration design assistant. Both represent the same \texttt{YAML} configuration. Users can place new objects in the environment with definable positions, rotations, sizes, and colours, and the assistant detects any overlapping objects. When a configuration contains such an overlap, the problematic items are spawned in separate locations according to the precedence set out in the configuration file. This is exemplified in the right panel, where a ramp is spawned in the bottom left portion of the arena, rather than to the right of the central blue wall as displayed in the configuration assistant representation. The assistant signals the potential overlap with a red border around the object. New items can effortlessly be added and supported by the assistant as the environment develops. The assistant is lightweight and quicker to launch than Animal-AI, thus streamlining configuration design. It has been used by by researchers to speed up the experimental design process.

Task configurations can also be procedurally generated. A simple accompanying command line tool is provided to design and generate precisely specified variations of a task.\footnote{See \url{https://github.com/wschella/procgen-companion}.} Each generated file is a valid Animal-AI configuration. The tool supports selecting random colours, specifying a fixed list of options, conditional values dependent on other choices already made, labels, and much more. Procedural generation allows for the quick creation of large test batteries, making behavioural evaluation more robust. For instance, researchers can study whether the performance of agents is sensitive to changes that ought to be irrelevant to success on the task, such as the colours of some objects. 

To demonstrate the ease with which one can design and run experiments in the Animal-AI Environment, we present three experiments. All coordinates are presented as ($x$,$y$,$z$) where $y$ is the vertical axis. The first experiment examines goal-directed navigation using a simple foraging task, emulating experiments of foraging behaviour in a wide variety of animals (see, e.g., \cite{pyke1977optimal}, \cite{schoener1971theory}). The agent is spawned at a random $x$-coordinate at ($x$,0,5), with a random initial rotation. At the centre of the arena (20,0,20) is a \texttt{SpawnerTree}, which spawns 10 \texttt{GoodGoalMulti} goals of size 2. The agent must obtain them all before their health runs out. The agent maximises reward by finding the shortest path between the spawned goals, known as \textit{foraging by energy maximisation} (\cite{king2022optimal, pyke1977optimal}).

\begin{figure}[h]
\centering
    \includegraphics[width=0.9\linewidth]{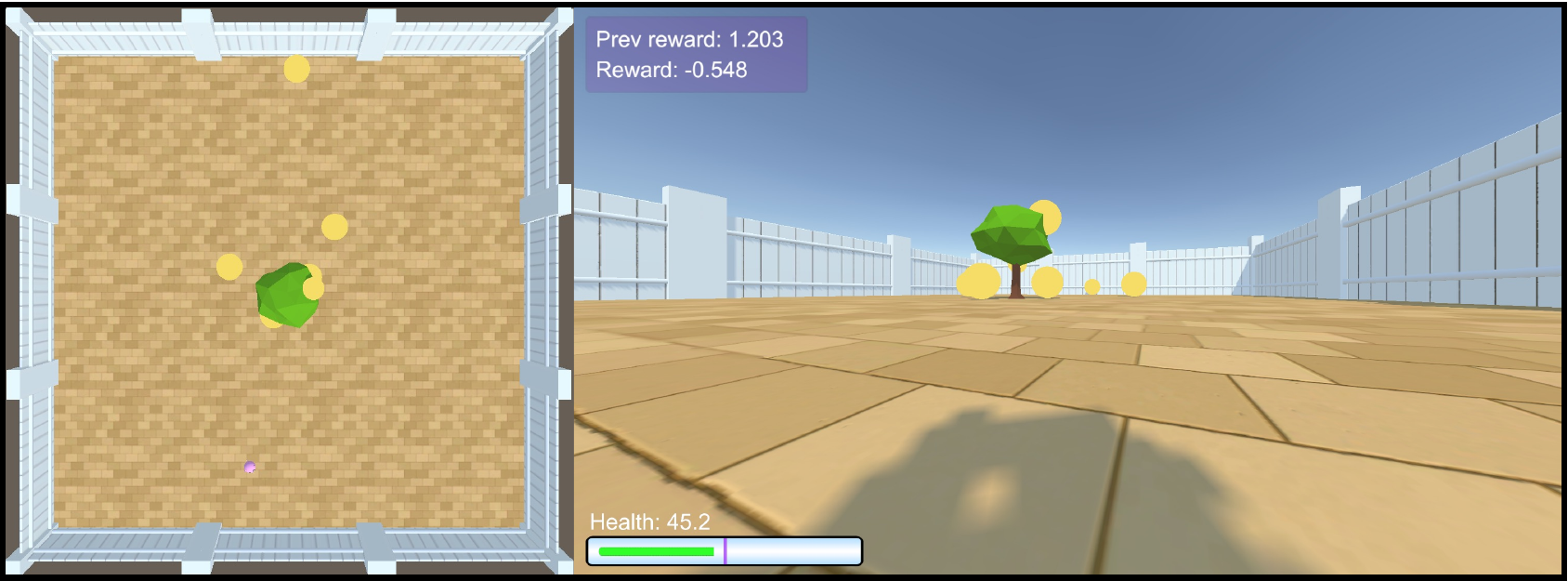}
    \caption{The foraging task experiment. \textbf{Left}: A bird's-eye view of the task. \textbf{Right}: The agent's viewpoint during the task.}
\label{fig:ForagingExperiment}
\end{figure}

The second experiment is based on the well-known \textit{operant chamber} experiments (\cite{heron1939apparatus}) that have been used to study learning in a wide variety of animals (\cite{macphail1982vertebrates}). In the classic design, the animal is placed in a chamber with a lever which can be pressed to return a food reward. Naive animals from which food has been withheld for a few hours are placed in the chamber, and learn through trial-and-error to operate the lever in return for rewards. An important feature of this task design in early work on animal behaviour was that lever presses were automatically recorded. Since the operant chamber is straightforward to design, several animals could be studied in parallel, a property mirrored by the Animal-AI Environment for computational models. We recreate a version of an operant chamber in the Animal-AI Environment by using the \texttt{SpawnerButton} object placed at (38.5,0,18.8), which, when pressed, spawns a \texttt{GoodGoal} next to it at (39,3,22). The button is enclosed by walls on three sides to ensure that the agent approaches it from the front. The agent is spawned at a random position in the arena. To succeed, the agent must press the button and then navigate to obtain the reward. We also construct a training curriculum, in line with work in reinforcement learning (\cite{narvekar2020curriculum}) and on shaping in animal behaviour theory (\cite{skinner1958reinforcement, peterson2004day}). The curriculum introduces the agent to the button and the consequences of pressing it. First, the agent is spawned at (36,0,20), directly in front of the bottom, such that it is likely to hit it by accident. Then, it is spawned further away, at (30,0,20), and then at (20,0,20). In the fourth stage, it is spawned at (20,0,$z$), where $z$ is randomly selected. Finally, it is presented with the final task, in which it is spawned at a random ($x$,$z$) coordinate. By comparing performance with and without the curriculum, its effect on learning can be directly assessed.

\begin{figure}[h]
\centering
    \includegraphics[width=0.9\linewidth]{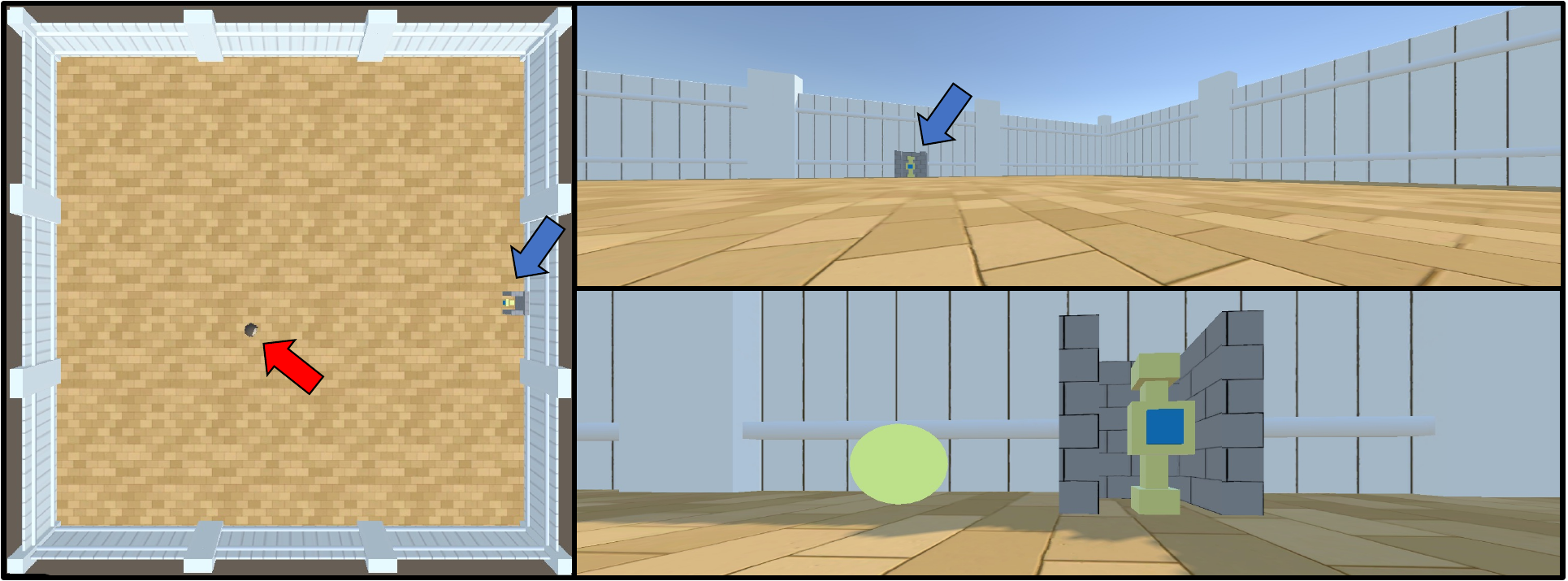}
    \caption{The Operant Chamber task. The agent is indicated by the red arrow and the location of the button is indicated by the blue arrows. \textbf{Left}: A bird's-eye view of the task. \textbf{Right}: The agent's viewpoint during the task.}
\label{fig:operantChamberTask}
\end{figure}

\begin{figure}[h]
    \centering
    \includegraphics[width=\linewidth]{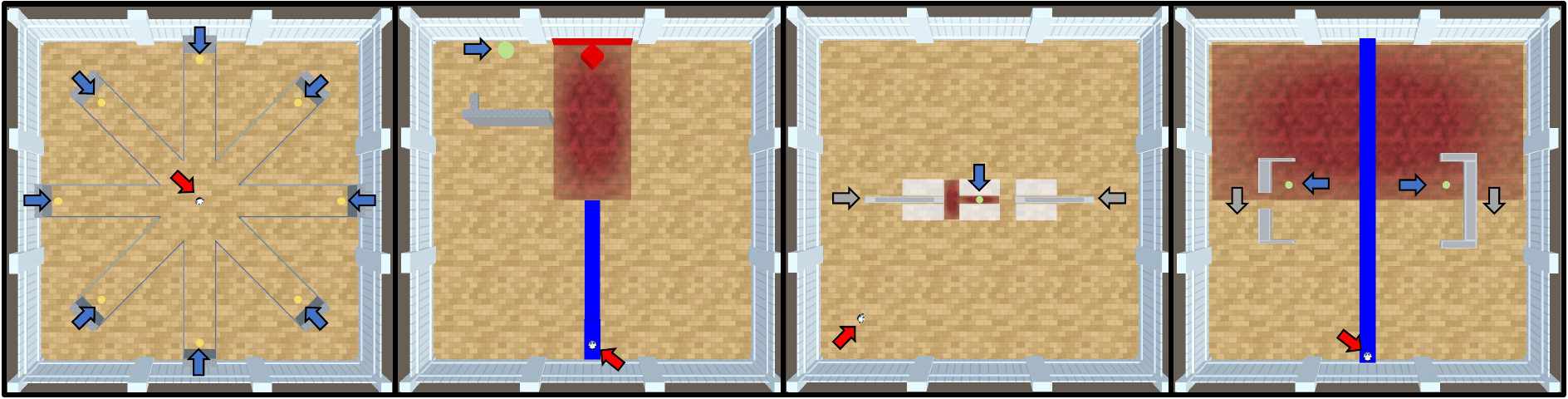}
    \caption{Four tasks from the Animal-AI Testbed. Red arrows indicate the position of the agent, blue arrows indicate the position of goals in the environment, and grey arrows indicate (possible) directions of motion of movable objects. \textbf{Far Left}: A Radial Arm Maze---the agent starts in the centre and must collect the eight \texttt{GoodGoalMulti} at the end of each arm. \textbf{Centre Left}: An object permanence task---the agent first observes a \texttt{GoodGoal} in the middle of a \texttt{DeathZone}, then the lights go out while the \texttt{GoodGoal} rolls behind a \texttt{Wall} on the left. \textbf{Centre Right}: A trap-tube task---pushing the block on the right would knock the \texttt{GoodGoal} into the \texttt{DeathZone}, while the left block would not. \textbf{Far Right}: A string-pulling task---only the \texttt{UBlock} can be used to pull the \texttt{GoodGoal} out of the \texttt{DeathZone}.}
    \label{fig:animal-ai-tasks}
\end{figure}

Both the foraging task and the button tasks are on the simple end of possible experiments that can be conducted in the Animal-AI Environment. For the third experiment, we provide the \textit{Animal-AI Testbed}, updated for compatibility with the latest version of the environment. The testbed is a series of 900 configurations inspired by comparative cognition research, used in the Animal-AI Olympics competition held in 2019 (see \cite{crosby2020animal,crosby2019animal,voudouris2022direct}). The Animal-AI Testbed is divided into 10 levels which probe different parts of spatial and physical cognition (see Table \ref{tab:olympics-test-bed}). Each level consists of 30 tasks with different demands and difficulty, each task of which has three variants, varying, for example, the starting position of the agent or the size of the goals. Examples of four tasks are presented in Figure \ref{fig:animal-ai-tasks}.

\begin{table}
\centering
 \caption{The Animal-AI Testbed consists of 900 configurations divided into 10 levels of 30 tasks with 3 variants each. Each level focuses on a different aspect of spatial and physical cognition (see \cite{voudouris2022direct} for more details).}
  \begin{tabular}{p{0.4\linewidth} | p{0.5\linewidth}}
    \toprule
    Level     & Description  \\
    \midrule
    \rule{0pt}{1ex}
     Food Retrieval (L1) & Valenced objects in an open arena with no obstacles. \\
    \rule{0pt}{3ex}
    Preferences (L2) & Valenced objects in forced- and free-choice configurations, such as Y-Mazes and Delayed Gratification tasks. \\
    \rule{0pt}{3ex}
    Static Obstacles (L3) & Valenced objects are fully or partially occluded by movable and immovable objects. \\
    \rule{0pt}{3ex}
    Avoidance (L4) & Positively valenced objects are arranged around negatively valenced objects. To obtain the former, the agent must navigate carefully to avoid the latter. \\
    \rule{0pt}{3ex}
    Spatial Reasoning and Support (L5) & Positively valenced objects are (partially) occluded. The agent must infer their presence based on their absence elsewhere. Valenced objects may also be supported out of reach by other objects, such as movable blocks.\\
    \rule{0pt}{3ex}
    Generalisation (L6) & Tasks from L1-L5 but adapted to have different features, such as the walls, floor, and/or ceiling being different colours. \\
    \rule{0pt}{3ex}
    Internal Modelling (L7) & Tasks from L1-L5 with different lights out intervals.\\
    \rule{0pt}{3ex}
    Object Permanence and Working Memory (L8) & Valenced objects become occluded behind other objects. \\
    \rule{0pt}{3ex}
    Numerosity and Advanced Preferences (L9) & Several valenced objects of different sizes and in different quantities are present in forced-choice settings. The agent must visually determine which choice will lead to the greatest reward.\\
    \rule{0pt}{3ex}
    Causal Reasoning (L10) & Positively valenced objects can only be obtained by interacting with neutral, movable objects, such as blocks.\\
    \bottomrule
  \end{tabular}
  \label{tab:olympics-test-bed}
\end{table}

\section{A Battery of Agents}\label{sec:battery-agents}

We provide implementations of three main agent types for running computational experiments in the Animal-AI Environment: \textit{Random Action Agent}, \textit{Heuristic Agent}, and \textit{Learner}. Each of these showcase the versatility of the environment for running experiments in multiple disciplines. Implementation of these agents is available in the supplemental material (see Section \ref{sec:supplemental}).

\paragraph{Random Action Agents}
When performing testing on artificial and biological agents, it is often useful to define \textit{chance performance} on tests that have been developed. However, experiments might be very complex, meaning that chance performance is not easily defined analytically. We provide customisable random action agents to simulate chance performance. The number of steps before picking a new action can be defined, or drawn from one of several distributions, and the agent can be biased to pick certain actions over others, or to correlate future actions with actions it has previously taken.

\paragraph{Heuristic Agents}
More sophisticated agents can be designed by hand-coding rules for action specification in response to different observations. This can be done using raycast observations or camera observations. We provide examples of how this logic can be set up, to enable sophisticated, but rigid, behaviours in the environment.

\paragraph{Learners}
The Animal-AI Environment is integrated with \texttt{Gymnasium}, as well as \texttt{stable-baselines-3} (\cite{stable-baselines3}), rendering available many common (deep) reinforcement learning algorithms, such as Proximal Policy Optimisation (PPO, \cite{schulman2017proximal}) and Deep Q-Learning (DQN, \cite{mnih2013playing}).
We also provide an integration of \texttt{Dreamer-v3}, a recent model-based reinforcement learning algorithm that achieved state-of-the-art performance in Minecraft (\cite{hafner2023mastering}).  

\subsection{Experiments}

Using the tests designed in Section \ref{sec:Designing Experiments in Animal-AI} as well as the Animal-AI Testbed (\cite{crosby2019animal, crosby2020animal}), we present the performances of a subset of the agents that have been implemented in the Animal-AI Environment.

\subsubsection{Foraging Experiment}\label{sec:foraging-experiment}

\begin{figure}[h]
\centering
    \includegraphics[width=1\linewidth]{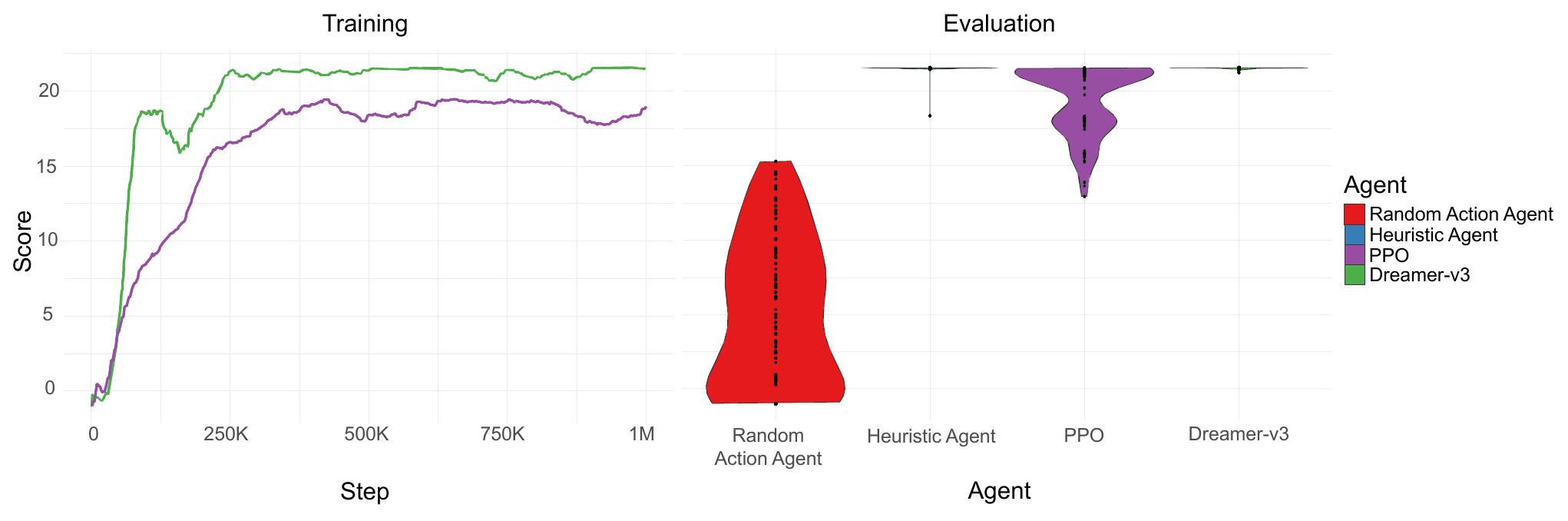}
    \caption{The results of four agents on the Foraging Task. Left: The training curves for PPO and Dreamer-v3 over one million training steps. Right: The scores of each type of agent over 100 trials of the task. The maximum reward available is 22, disregarding the reward lost during navigation between goals.}
\label{fig:ForagingBoxPlots}
\end{figure}

We studied four types of agents on the foraging task in Figure \ref{fig:ForagingExperiment}. To serve as a baseline for chance performance on this test, we used a random action agent. This agent selects one of 9 actions with equal probability. The number of steps to take performing that action is a positive number drawn from a clipped normal distribution with mean of 5 and a standard deviation of 1, rounded to the nearest integer. We also designed a heuristic agent, to provide an example of near-optimal performance on this task. This agent navigates towards good goals using raycasts (45 rays over the frontal 60 degrees). When it gets stuck (i.e., its velocity is near 0 in all dimensions), it moves forwards and left or forwards and right with a probability of 0.5 until its velocity increases again. We then trained two deep reinforcement learning agents, Proximal Policy Optimisation (PPO) and Dreamer-v3, on the task, with 64$\times$64 pixel input for one million steps each. We used the XL variant of Dreamer-v3, which has 200M learnable parameters. We then evaluated all four agents on 100 episodes of the Foraging task. The training curves for PPO and Dreamer-v3, as well as the distribution of results for all four agents, are presented in Figure \ref{fig:ForagingBoxPlots}. Note that all training curves presented in this paper are smoothed using 100-episode partial moving averages. Dreamer-v3, PPO, and the Heuristic Agent performed similarly and above the level of chance defined by the random action agent, evidence of efficient foraging strategies that approximately maximise reward. A Kruskall-Wallis Rank Sum Test indicates that there was a significant difference between the agents ($\chi^2$(3) = 318.16, p < 0.00001). Post-hoc Dunn Test statistics (\cite{FSADunnTest}) indicate that all agents are significantly different from each other, except for Dreamer-v3 and the Heuristic Agent (see Appendix \ref{app:further-results-foraging}).

\subsubsection{Operant Chamber Experiment}\label{sec:button-press-experiment}

We studied four types of agents on the Operant Chamber task in Figure \ref{fig:operantChamberTask}. As with the previous experiment, we used a random action agent to define chance performance. A heuristic agent was also designed. This agent navigates towards \texttt{SpawnerButton}, \texttt{GoodGoal} and immovable objects using raycasts (45 rays over the frontal 60 degrees). When it has not detected any objects it repeats the previous action. When it is stationary, it moves forwards and left. A notable limitation is that raycasts do not detect which side of the button needs to be pressed, so this agent can successfully navigate towards buttons, but it does not necessarily press the button. Therefore, vision-based agents will likely perform much better, as the button is indicated by colour only.

\begin{figure}[h]
\centering
    \includegraphics[width=1\linewidth]{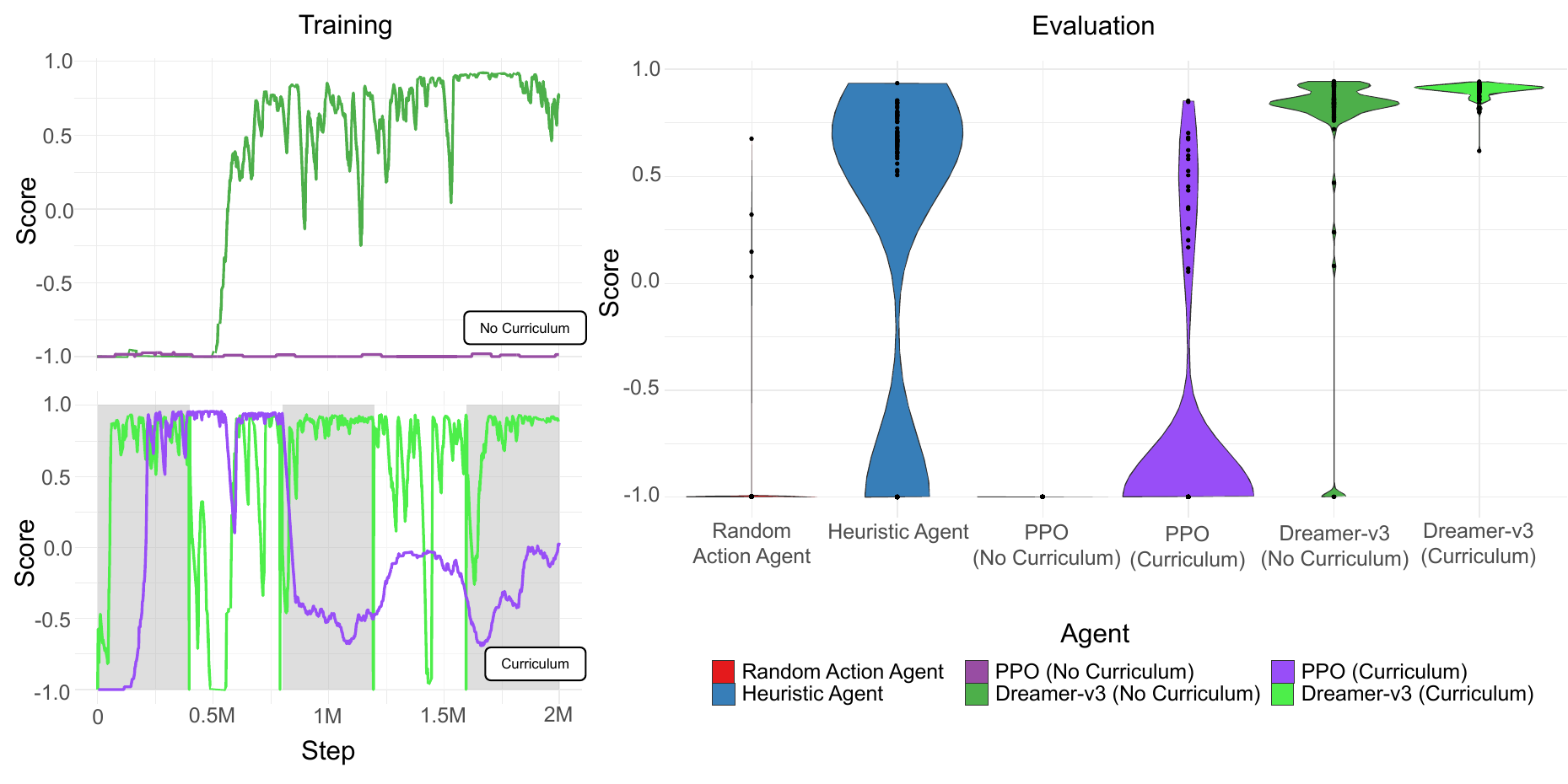}
    \caption{The results of six agents on the Operant Chamber Task. Left: The training curves for PPO and Dreamer-v3 over two million training steps with (bottom) and without (top) a training curriculum. Alternating grey and white panels indicate stages of the curriculum. Right: The scores of each agent over 100 trials of the task used in the \textit{No Curriculum} training condition. The maximum reward available is 1, disregarding the reward lost during navigation.}
\label{fig:operantChamberPlots}
\end{figure}

As with the previous experiment, we trained PPO and Dreamer-v3 on the task with 64$\times$64 pixel input for two million steps each. An agent was trained with and without the curriculum. The curriculum was cumulative, with the agents trained on 5 stages; each stage consisted of on an Animal-AI arena configuration containing the current stage of the curriculum along with all the previous stages interleaved. The agents loop over the full multi-arena configuration for 400,000 steps, before moving to the next stage, leading to training over two million steps in total. We then evaluated all six agents on 100 episodes of the Foraging task. The training curves for the four PPO and Dreamer-v3 agents, as well as the distribution of results for all six agents, are presented in Figure \ref{fig:operantChamberPlots}. The Heuristic Agent performed above the level of chance defined by the random action agent, as did both Dreamer-v3 agents, with the agent trained on the curriculum performing better. The PPO agent trained only on the Operant Chamber Task did not learn to press the button. With the curriculum, PPO was able to learn to solve some configurations, again demonstrating the role of curricula in scaffolding learning in these agents. A Kruskall-Wallis Rank Sum Test indicates that there was a significant difference between the agents ($\chi^2$(3) = 400.15, p < 0.00001). Post-hoc Dunn Test statistics indicate that all agents are significantly different from each other, except for the Heuristic Agent and both PPO agents compared to the random action agent, and the PPO agents compared to each other, suggesting that these agents are performing no better than chance. Notably, the curriculum led to significantly better performance for Dreamer-v3 (\textit{Bonferroni}-adjusted $p=$0.035504; see Appendix \ref{app:further-results-operant}).

\subsubsection{The Animal-AI Testbed}\label{sec:aai-testbed-experiment}

While many agents in the Animal-AI Olympics were able to solve individual tasks in the Animal-AI Testbed, performance was low compared to humans. Children aged 6-10 performed significantly better than even the top-performing AI systems in the competition (\cite{voudouris2022direct}).

Here, we present the performances of six agents on the Animal-AI Testbed, along with data from children aged 6-10. As with the foraging experiment (Section \ref{sec:foraging-experiment}), we used a random action agent to define chance performance, and a heuristic agent with 45 rays emitted over the frontal 60 degrees of its view. We then trained PPO and Dreamer-v3 on 64$\times$64 pixel input on a cumulative curriculum to be described below. We evaluated these four agents on all three variants of the 300 tasks in the Animal-AI Testbed (n=900). As far as we know, this is the first example of agents being trained and tested on the Animal-AI Testbed (although see \cite{lind2021can} for a variation).

We also present below the performances of children (n=59) aged 6-10 on a subset of the tasks as a reference, taken from openly accessible data from Voudouris et al. (\citeyear{voudouris2022direct}).\footnote{Children played a randomly selected set of 4 tasks per level. They played these tasks randomly within each level, but with the levels arranged in the order presented in Table \ref{tab:olympics-test-bed}. Full data are available at \href{https://osf.io/g8u26/}{https://osf.io/g8u26/}.} We also present performances from the two best performing agents from the Animal-AI Olympics competition, \textit{Trrrrr} and \textit{ironbar}, for reference (\cite{crosby2019animal,crosby2020animal}). \textit{Trrrrr} is a customised PPO architecture trained on 84$\times$84 RGB pixel input on a bespoke curriculum for 70M steps and achieved the highest overall score in the competition.\footnote{See \href{https://github.com/Denys88/rl_animal}{https://github.com/Denys88/rl\_animal}.} \textit{ironbar} is a customised PPO architecture trained on 84$\times$84 RGB pixel input on a bespoke curriculum for 20M steps and achieved the second highest overall score in the competition.\footnote{See \href{https://github.com/ironbar/orangutan}{https://github.com/ironbar/orangutan}.} The Testbed was not released to the developers of these agents until after their agents had been developed, trained, and tested.

PPO and Dreamer-v3 were trained on a cumulative curriculum consisting of 11 stages. The first stage contained one variant from each of 30 tasks in \textit{Food Retrieval} (L1), totalling 30 configurations which the agent looped over for two million steps. The second stage contained one variant from each of the 60 tasks in L1 and \textit{Preferences} (L2), totalling 60 configurations. The pattern repeats until the tenth stage, containing one variant from each of the 300 tasks in L1-L10. The agents were trained for two million steps on the first ten stages. They were then trained for a further five million steps on the last stage again (300 configurations from tasks L1-10), to ensure adequate experience with all the tasks in the Testbed. Agents were therefore trained for a total of 25 million steps.

\begin{figure}[h]
\centering
    \includegraphics[width=1\linewidth]{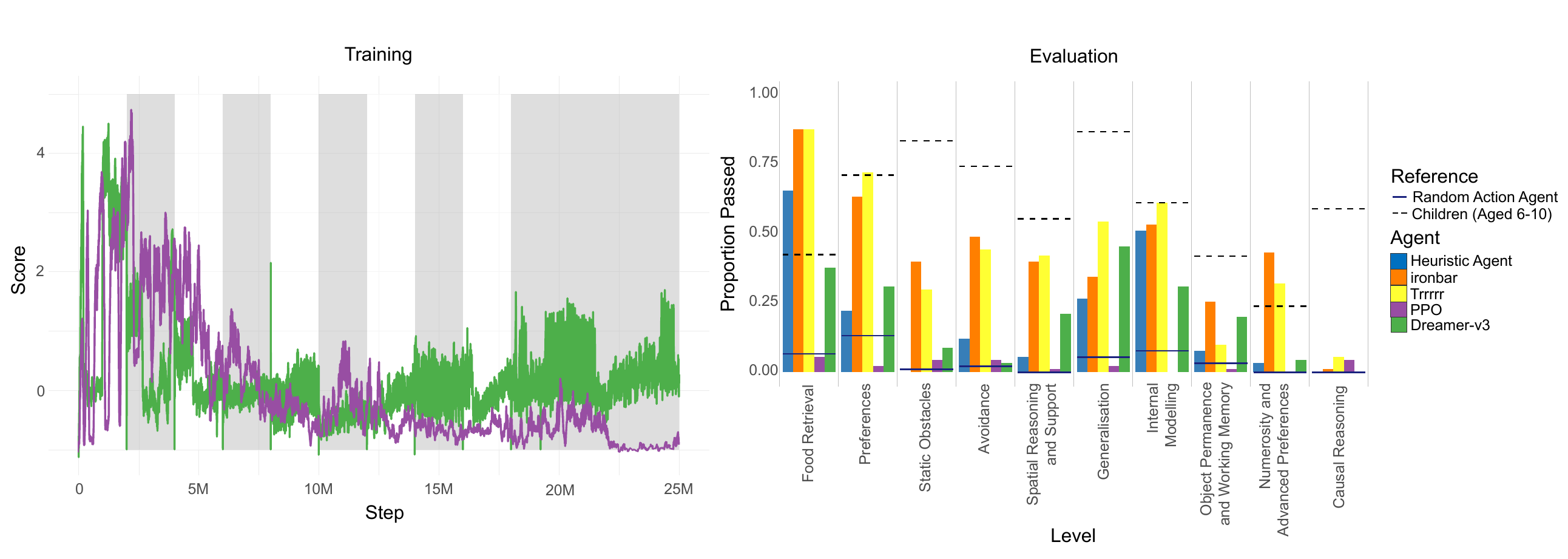}
    \caption{The results of seven types of agent on the Animal-AI Testbed. Left: The training curves for PPO and Dreamer-v3 over 25 million training steps. Alternating grey and white panels indicate stages of the curriculum. Right: The proportion of Animal-AI Testbed tasks that each agent passed, according to the level. Data are presented for each agent with one score per episode. Chance performance is defined by the Random Action Agent. Human performance is aggregated over data from 6-10 year olds (n=59) on a subset of four tasks per level.}
\label{fig:competitionBarChart}
\end{figure}

Figure~\ref{fig:competitionBarChart} presents the proportion of tests passed by each agent on each level, with the proportion for random action agents and for children indicated by reference lines. \textit{Trrrrr} and \textit{ironbar} performed the best across the levels, performing comparably on \textit{Preferences} and exceeding children on \textit{Food Retrieval}. The heuristic agent also performed competitively on \textit{Food Retrieval} and \textit{Internal Modelling}. Dreamer was generally outperformed by these other agents, but performed above the chance level defined by the sample of random action agents on \textit{Food Retrieval}, \textit{Preferences}, \textit{Generalisation}, and \textit{Internal Modelling}. PPO performed poorly across the Testbed, generally performing at or near to chance performance, on most levels. Overall, while \textit{Trrrrr} and \textit{ironbar} placed first and second respectively in the Animal-AI Olympics Competition, Dreamer-v3 would have placed 25\textsuperscript{th}, the Heuristic Agent would have placed 27\textsuperscript{th}, the Random Action Agent would have placed 60\textsuperscript{th}, and PPO would have placed 61\textsuperscript{st} out of 64.

Three levels (Causal Reasoning, Object Permanence and Working Memory, and Numerosity and Advanced Preferences) pose difficulty for all agents, including children, indicating that the Animal-AI Testbed remains a worthy challenge for state-of-the-art deep reinforcement learning. Although both Dreamer-v3 and PPO were trained on a curriculum of increasing difficulty, we leave it to future work to determine whether an alternative curriculum, or an adaptively generated curriculum (e.g., \cite{humantimescaleadaptation2022}), would be more suitable for handling the diversity and complexity of the Animal-AI Testbed. See Section \ref{app:further-results-aai_t} for further analysis and results. While better performance could be achieved through more training and better curricula, we leave these investigations to future work.

\section{Conclusion}

In this paper, we have presented the Animal-AI Environment, a virtual laboratory to bring together researchers from comparative cognition and artificial intelligence. Future iterations will continue to build on the Animal-AI Environment to improve its utility for researchers in these fields. Areas of active improvement include \textit{adjustable arena sizes}, allowing researchers to build, for example, open-field navigation experiments. The introduction of \textit{arena themes}, enabling researchers to replace the walls, floor, and sky with alternative colours and textures, will enable more naturalistic replication of experiments from the physical laboratory. We will also continue to improve the Animal-AI Environment's integration with cutting-edge frameworks and pipelines in artificial intelligence and machine learning, to ensure that the latest computational models are available for use.

The Animal-AI environment is unique in its mission to be a common tool for research on animal behaviour and cognition, and its integration with computational modelling and artificial intelligence. By bringing the precision of computational models of the brain and behaviour in touch with the carefully constructed experimental designs, datasets, and sophisticated cognitive theories, we can make significant progress towards better understanding animal cognition and behaviour.

\newpage

\section*{Declarations}

\subsection*{Funding}

This work was funded by an ESRC DTP scholarship (ES/P000738/1), the Kinds of Intelligence Program at the Leverhulme Centre for the Future of Intelligence (G108086), the US DARPA HR00112120007 (RECoG-AI) grant, and a Templeton World Charity Foundation grant (Major Transitions in the Evolution of Cognition: TWCF-2020-20539). This work was also funded by the European Union with the ``Programa Operativo del Fondo Europeo de Desarrollo Regional (FEDER) de la Comunitat Valenciana 2014-2020'' under agreement INNEST/2021/317 (Neurocalçat) and by the Vic. Inv. of the Universitat Politècnica de València under ``programa de ayudas a la formación de doctores en colaboración con empresas'' (DOCEMPR21).

\subsection*{Conflicts of Interest}

M. Crosby is now employed by DeepMind Technologies Limited, but this work was conducted prior to this employment. The remaining authors declare that the research was conducted in the absence of any commercial or financial relationships that could be construed as a potential conflict of interest.

\subsection*{Ethics Approval}

Ethical approval is not required because no experiments were performed with humans or animals. All data from human children were from an openly accessible repository (https://osf.io/g8u26/) associated with an independent study that received ethical approval.

\subsection*{Consent to Participate}

No research was conducted with human subjects. All data from human children were from an openly accessible repository (\href{https://osf.io/g8u26/}{https://osf.io/g8u26/}) associated with an independent study that received ethical approval.

\subsection*{Consent for Publication}

No research was conducted with human subjects. All data from human children were from an openly accessible repository (\href{https://osf.io/g8u26/}{https://osf.io/g8u26/}) associated with an independent study that received ethical approval.

\subsection*{Code Availability}\label{sec:supplemental}

Further information, tutorials, scripts, and documentation for the project can be found here: \href{https://github.com/Kinds-of-Intelligence-CFI/animal-ai}{https://github.com/Kinds-of-Intelligence-CFI/animal-ai}. Scripts to replicate the experiments and analyses in this paper can be found here: \href{https://github.com/Kinds-of-Intelligence-CFI/aai3-paper-experiments}{https://github.com/Kinds-of-Intelligence-CFI/aai3-paper-experiments}. 

\subsection*{Author Contributions}

\textbf{K. Voudouris}: Conceptualization, Methodology, Software, Validation, Formal Analysis, Investigation, Resources, Data Curation, Writing (Original Draft), Writing (Review \& Editing), Visualization, Project Administration. \textbf{I. Alhas}: Software, Validation, Data Curation, Writing (Review \& Editing). \textbf{W. Schellaert}: Software, Validation, Investigation, Data Curation, Writing (Review \& Editing). \textbf{M. G. Mecattaf}: Formal Analysis, Software, Validation, Investigation, Data Curation, Visualization, Writing (Review \& Editing). \textbf{B. Slater}: Software, Validation, Writing (Review \& Editing). \textbf{M. Crosby}: Conceptualization, Formal Analysis, Software, Writing (Review \& Editing). \textbf{J. Holmes}: Software, Writing (Review \& Editing). \textbf{J. Burden}: Formal Analysis, Writing (Review \& Editing). \textbf{N. Chaubey}: Formal Analysis, Writing (Review \& Editing). \textbf{N. Donnelly}: Data Curation, Writing (Review \& Editing). \textbf{M. Patel}: Software, Data Curation, Writing (Review \& Editing). \textbf{M. Halina}: Conceptualization, Resources, Supervision, Funding Acquisition, Writing (Review \& Editing). \textbf{J. Hern\'{a}ndez-Orallo}: Conceptualization, Resources, Supervision, Funding Acquisition, Writing (Review \& Editing). \textbf{L. G. Cheke}: Conceptualization, Methodology, Resources, Supervision, Funding Acquisition, Writing (Review \& Editing).

\subsection*{Availability of Data and Materials}\label{sec:supplemental}

All datasets and materials for the experiments reported here can be found at \href{https://github.com/Kinds-of-Intelligence-CFI/aai3-paper-experiments}{https://github.com/Kinds-of-Intelligence-CFI/aai3-paper-experiments}. None of the experiments reported here were preregistered.

\newpage

\printbibliography

\newpage

\appendix

\section{Sources of Maintenance Statuses of Other Environments}\label{app:maintenance-status}

We defined maintenance status according to the last commit on the respective GitHub repository for each environment, as of 17\textsuperscript{th} January 2025. Further information is presented in Table \ref{tab:maintenance-dates}.

\begin{table}[h]
\centering
 \caption{The dates of the last commit in the GitHub history of each environment presented in Table \ref{tab:competitor-overview}. AAI3 is omitted as it is under constant active development across several repositories. These data are correct as of 2025-Jan-17.}
  \begin{tabular}{ p{0.13\linewidth}p{0.2\linewidth} p{0.6\linewidth}}
    \toprule
    Environment & 
    Date of Last Commit & Source \tabularnewline
    \midrule
    DM Lab & 2023-Jan-04 & \href{https://github.com/google-deepmind/lab/commits/master/}{https://github.com/google-deepmind/lab/commits/master/}\tabularnewline
    XLand & \textbf{NA} & \textbf{No Public Release} \tabularnewline
    Avalon & 2023-May-04 & \href{https://github.com/Avalon-Benchmark/avalon/commits/main/}{https://github.com/Avalon-Benchmark/avalon/commits/main/}\tabularnewline
    MuJoCo & 2025-Jan-16 & \href{https://github.com/google-deepmind/mujoco/commits/main/}{https://github.com/google-deepmind/mujoco/commits/main/}\tabularnewline
    MineRL & 2024-Nov-26 & \href{https://github.com/minerllabs/minerl/commits/dev/}{https://github.com/minerllabs/minerl/commits/dev/}\tabularnewline
    Malm\"{o} & 2022-Sep-23 & \href{https://github.com/microsoft/malmo/commits/master/}{https://github.com/microsoft/malmo/commits/master/}\tabularnewline
    MineDoJo & 2023-Aug-29 & \href{https://github.com/MineDojo/MineDojo/commits/main/}{https://github.com/MineDojo/MineDojo/commits/main/}\tabularnewline
    ThreeDWorld & 2024-May-31 & \href{https://github.com/threedworld-mit/tdw/commits/master/}{https://github.com/threedworld-mit/tdw/commits/master/}\tabularnewline
    MiniWorld & 2025-Jan-12 & \href{https://github.com/Farama-Foundation/Miniworld/commits/master/}{https://github.com/Farama-Foundation/Miniworld/commits/master/}\tabularnewline
    APExplorer 3D & \textbf{NA} & \textbf{No Public Release} (a static version linked to \citeauthor{allritz2022chimpanzees} (\citeyear{allritz2022chimpanzees}) is available in the supplementary material)  \tabularnewline
    \bottomrule
  \end{tabular}
  \label{tab:maintenance-dates}
\end{table}

\section{Further Experimental Results}\label{app: further-results}

\subsection{Foraging Experiment}\label{app:further-results-foraging}

See Table \ref{tab:dunn-test-foraging} for post-hoc Dunn Test statistics for performances on the foraging task.

\begin{table}[h]
\centering
 \caption{Post-Hoc Dunn Test of Multiple Comparisons for performances on the foraging task. Unadjusted and Bonferroni-Adjusted $p$-values are presented. All values are presented to 6 decimal places.}
  \begin{tabular}{lrrr}
    \toprule
    Comparison     & Z-statistic     & Unadjusted $p$-value & Adjusted $p$-value \\
    \midrule
    Random - Heuristic & 14.324042  & <0.000001 & <0.000001     \\
    Random - PPO & 6.664166 & <0.000001 & <0.000001      \\
    Random - Dreamer  & 15.662257  & <0.000001 & <0.000001  \\
    Heuristic - PPO &  7.659876 & 0.001626 & 0.009759 \\
    Heuristic - Dreamer & 1.338215 & 0.180826 & 1.000000 \\
    PPO - Dreamer & 8.998092 & <0.000001 & <0.000001 \\
    \bottomrule
  \end{tabular}
  \label{tab:dunn-test-foraging}
\end{table}

\subsection{Operant Chamber Experiment}\label{app:further-results-operant}

See Table \ref{tab:dunn-test-operant} for post-hoc Dunn Test statistics for performances on the Operant Chamber Task.

\begin{table}[h]
\centering
 \caption{Post-Hoc Dunn Test of Multiple Comparisons for performances on the Operant Chamber Task. Unadjusted and Bonferroni-Adjusted $p$-values are presented. All values are presented to 6 decimal places.}
  \begin{tabular}{lrrr}
    \toprule
    Comparison     & Z-statistic     & Unadjusted $p$-value & Adjusted $p$-value \\
    \midrule
    Random - Heuristic & 2.258691  & 0.023903 & 0.358539    \\
    Random - PPO (No Curriculum) & -2.314548 & 0.020638 & 0.309565      \\
    Random - PPO (Curriculum) & -0.779929 & 0.435433 & 1.000000      \\
    Random - Dreamer (No Curriculum)  & 10.014743  & <0.000001 & <0.000001  \\
    Random - Dreamer (Curriculum)  & 13.054605 & <0.000001 & <0.000001  \\
    Heuristic - PPO (No Curriculum) &  4.573239 & 0.00005 & 0.000072 \\
    Heuristic - PPO (Curriculum) &  3.038620 & 0.002377 & 0.035650 \\
    Heuristic - Dreamer (No Curriculum) & 7.756053 & <0.000001 & <0.000001 \\
    Heuristic - Dreamer (Curriculum) & 10.795914 & <0.000001 & <0.000001 \\
    PPO (No Curriculum) - Dreamer (No Curriculum) & 12.329291 & <0.000001 & <0.000001 \\
    PPO (No Curriculum) - Dreamer (Curriculum) & 15.369153 & <0.000001 & <0.000001 \\
    PPO (No Curriculum) - PPO (Curriculum) & -1.534619 & 0.124877 & 1.000000 \\
    PPO (Curriculum) - Dreamer (No Curriculum) & 10.794672&  <0.000001 & <0.000001 \\
    PPO (Curriculum) - Dreamer (Curriculum) & 13.834534 & <0.000001 & <0.000001 \\
    Dreamer (No Curriculum) - Dreamer (Curriculum) & -3.039861 & 0.002367 & 0.035503 \\
    \bottomrule
  \end{tabular}
  \label{tab:dunn-test-operant}
\end{table}

\subsection{The Animal-AI Testbed}\label{app:further-results-aai_t}

To examine the differences between agents and between levels on the Animal-AI Testbed, we used a generalised linear mixed effects model, implemented in \textit{R} (\cite{bates2015lme4}). Level and Agent are taken as fixed effects, with participant ID (either the name of the agent or the ID of the human participant) as a random intercept. Pass/fail is the response variable. The model, with a logit link function, was fitted by maximum likelihood (Adaptive Gauss-Hermite Quadrature) with the BOBYQA optimizer and 10 integration points.

In Table \ref{tab:odds-ratio-competition}, the odds ratios of success to failure are presented. The population of agents is significantly more likely to pass the \textit{Preferences} level compared to the first \textit{Food Retrieval} level, perhaps because the arena is more restricted with walls, so there is less space for exploration. They are also significantly more likely to pass the \textit{Generalisation} level. They are significantly more likely to fail \textit{Spatial Reasoning \& Support}, \textit{Object Permanence \& Working Memory}, \textit{Numerosity \& Advanced Preferences}, and \textit{Causal Reasoning}, indicating that these levels are difficult.

Taking the random action agent to define chance performance, PPO is not significantly more likely to pass the tasks than chance. Dreamer is significantly more likely to pass than chance, as were ironbar, Trrrrr, the Heuristic Agent, and children. Children are 47 times more likely to pass than chance, while ironbar and Trrrrr are 19 times more likely. Dreamer is six times more likely.

\begin{table}[ht]
\centering
 \caption{Odds ratios from a generalised linear mixed effects model. The reference category for \textit{Level} is \textit{Food Retrieval (L1)}, and for \textit{Agent} it is \textit{Random Action Agent}. All values are presented to 6 decimal places.}
  \begin{tabular}{llrr}
    \toprule
    Type & Fixed Effect & Estimated Odds   & $p$-value \\
    \midrule
    & Intercept & 0.048290 & 0.000003 \\
    \midrule
    Level & Preferences    & 1.383071     & 0.002403\\
    & Static Obstacles    & 1.046102     & 0.671138 \\
    & Avoidance    & 0.956010     & 0.671425 \\
    & Spatial Reasoning \& Support    & 0.562533     & <0.000001 \\
    & Generalisation    & 1.769805     & <0.000001 \\
    & Internal Modelling    & 1.057973     & 0.595556 \\
    & Object Permanence \& Working Memory    & 0.303834     & <0.000001 \\
    & Numerosity \& Advanced Preferences    & 0.184259     & <0.000001 \\
    & Causal Reasoning    & 0.367255     & <0.000001 \\
    \midrule
    Agent & Dreamer-v3 & 6.539571 & 0.038181 \\
    & PPO    & 0.622401     & 0.608951  \\
    & ironbar    & 21.925737     & 0.000644  \\
    & Trrrrr    & 22.146038     & 0.000617  \\
    & Heuristic Agent    & 6.164322    & 0.044706  \\
    & Children (Aged 6-10)   & 47.198110     & <0.000001 \\
    \bottomrule
  \end{tabular}
  \label{tab:odds-ratio-competition}
\end{table}

\end{document}